\documentclass[10pt,twocolumn,letterpaper]{article}

\usepackage{iccv}
\usepackage{times}
\usepackage{epsfig}
\usepackage{graphicx}
\usepackage{amsmath}
\usepackage{amssymb}
\usepackage{arydshln}
\usepackage[bottom]{footmisc}
\usepackage[accsupp]{axessibility}

\DeclareMathOperator*{\argmin}{arg\,min}

\usepackage[pagebackref=true,breaklinks=true,letterpaper=true,colorlinks,bookmarks=false]{hyperref}
\usepackage{xcolor}
\usepackage{multirow}
\iccvfinalcopy 

\def\iccvPaperID{10265} 

\newcommand{\parsa}[1]{}

\ificcvfinal\fi

\begin{document}

\title{PRANC: Pseudo RAndom Networks for Compacting deep models}

\author{Parsa Nooralinejad  \\
 {\small University of California, Davis}\\
 \vspace{-.1in}
\and
Ali Abbasi\\
 {\small Vanderbilt University}\\
 \vspace{-.1in}
\and
Soroush Abbasi Koohpayegani$\footnotemark[1]$\\
 {\small University of California, Davis}\\
 \vspace{-.1in}
\and
$\quad$$\quad$$\quad$$\quad$$\quad$Kossar Pourahmadi Meibodi $\footnotemark[1]$\\
 $\quad$$\quad$$\quad$$\quad$$\quad${\small University of California, Davis}\\
 \vspace{-.1in}
\and
Rana Muhammad Shahroz Khan$\thanks{Equal contribution}$$\quad$$\quad$$\quad$$\quad$$\quad$\\
 {\small Vanderbilt University}$\quad$$\quad$$\quad$$\quad$$\quad$\\
 \vspace{-.1in}
\and
Soheil Kolouri\\
 {\small Vanderbilt University}
\and
Hamed Pirsiavash\\
 {\small University of California, Davis}\\
}

\maketitle

\ificcvfinal\thispagestyle{empty}\fi
\begin{abstract}
We demonstrate that a deep model can be reparametrized as a linear combination of several randomly initialized and frozen deep models in the weight space. During training, we seek local minima that reside within the subspace spanned by these random models (i.e., `basis' networks). 
Our framework, PRANC, enables significant compaction of a deep model. The model can be reconstructed using a single scalar `seed,' employed to generate the pseudo-random `basis' networks, together with the learned linear mixture coefficients.
%
In practical applications, PRANC addresses the challenge of efficiently storing and communicating deep models, a common bottleneck in several scenarios, including multi-agent learning, continual learners, federated systems, and edge devices, among others. In this study, we employ PRANC to condense image classification models and compress images by compacting their associated implicit neural networks. PRANC outperforms baselines with a large margin on image classification when compressing a deep model almost $100$ times. Moreover, we show that PRANC enables memory-efficient inference by generating layer-wise weights on the fly. The source code of PRANC is here: \url{https://github.com/UCDvision/PRANC}

\end{abstract}

\section{Introduction}
\label{sec:intro}

The prevailing notion is that larger deep models yield improved accuracy. Yet, it remains unclear if the better generalization of larger models stems from the increased complexity of the architecture or more parameters. Moreover, among numerous good local minima in the loss function, training finds one. In this paper, we introduce a fresh approach: viewing a deep model as a linear combination within the weight space of several randomly initialized and frozen models. During learning, our goal shifts to finding a minimum that exists within the subspace defined by these initial models. Our findings highlight the potential to significantly compact deep models by retaining only the seed value of the pseudo-random generator and the coefficients for weight combination.

This efficient reparameterization benefits AI and ML applications by reducing deep model size for easier storage or communication. In modern neural networks with millions to billions of parameters, storage, and communication become costly. This issue worsens in low-bitrate environments due to physical constraints or adversarial disruption. For instance, underwater applications might have as low as $100$ bits per second bandwidth, then, transferring ResNet18's $11$M parameter model takes more than 40 days in such conditions. Moreover, in distributed learning with many agents, high-bandwidth WiFi networks still face congestion issues.

Going beyond communications, loading or storing these large models on edge devices poses another significant challenge. Edge devices often come with small memories unsuitable for storing large neural networks and may want to run the model less frequently (on-demand). Hence, they may benefit from compacting a deep model to fewer parameters to construct the model layer-by-layer or even kernel-by-kernel on-demand to run each inference. This will result in significantly less I/O cost.



\begin{figure}[t]
  \centering
  \includegraphics[width=\linewidth]{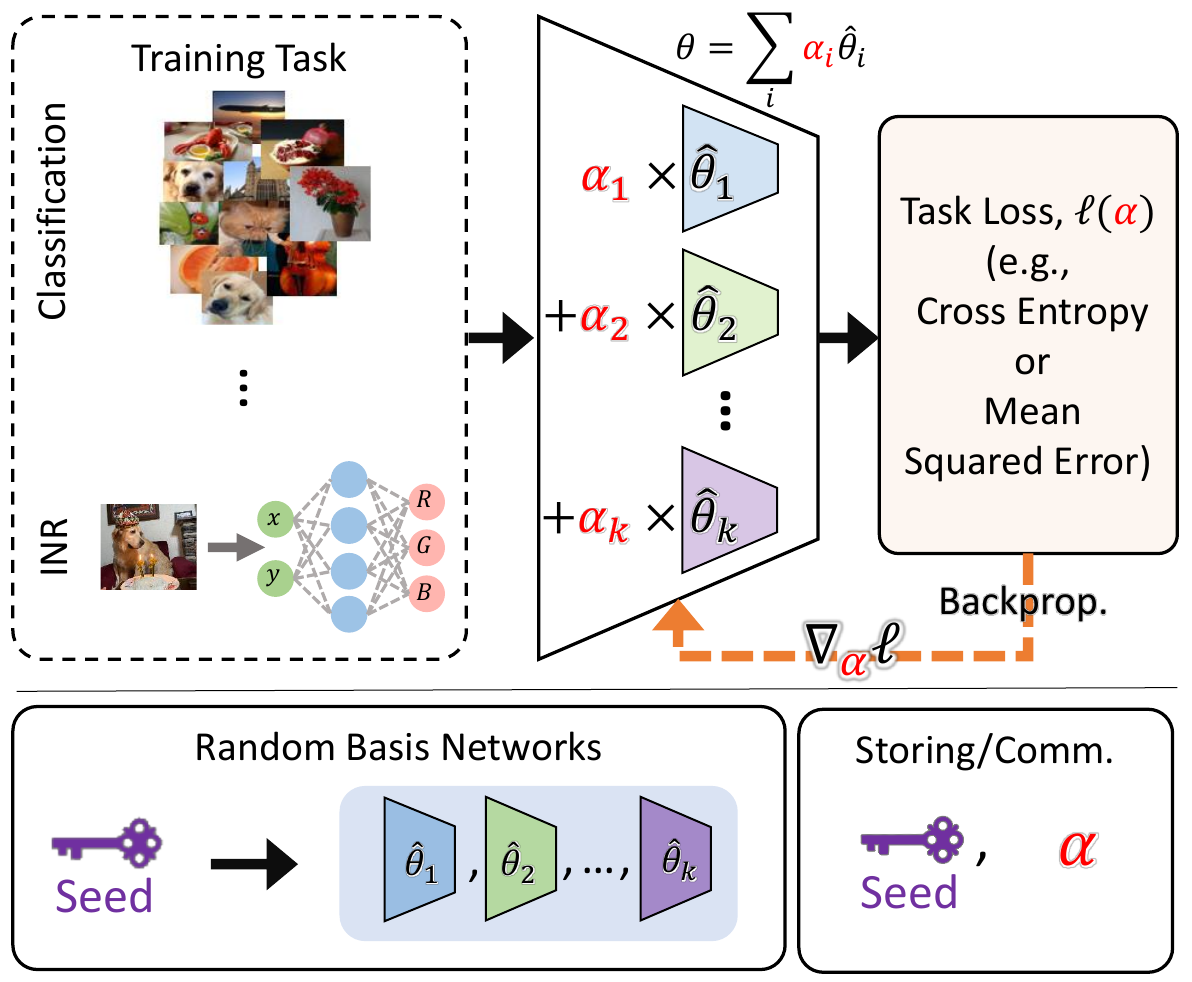}
  \caption{We restrict the deep model to be a linear combination of $k$ randomly initialized models. Since the number of models is much less than the size of the model, it is much less expensive to communicate or store the coefficients compared to the model or data itself. We tune $\alpha$ to minimize the loss of the task using standard backpropagation.
    \vspace{-.15in}
}
  \label{teaser}
\end{figure}

One may compact the model by distilling it into a smaller model \cite{hinton2015distilling}, pruning the model parameters \cite{lin2020dynamic}, quantizing the parameters \cite{lee2021network}, or sharing the weights as much as possible \cite{subia2022weight, chen2015compressing}. More recently, dataset distillation \cite{dataset_distil} is proposed. It can be seen as an alternative to model compression since one can store or communicate the distilled dataset and then train the model again when needed.
However, most of these methods are limited to small reduction factors, \eg, less than $30\times$. Also, knowledge distillation methods reduce the model architecture to a smaller one with fewer layers, which may limit the future application of that model, \eg, for future fine-tuning or lifelong learning, which needs the deeper architecture. 

We are interested in compacting a deep model by a considerable factor (\eg, $100\times$) without changing its architecture. The core idea behind our approach is simple. We constrain our model to be a linear combination of a finite set of randomly initialized models, called {\em basis} models. Hence, the problem boils down to finding the optimal linear mixture coefficients that result in a network that can solve the task effectively. The model can then be succinctly represented by the seed (a single scalar) to generate the pseudo-random basis models and the linear mixture coefficients (See Figure \ref{teaser}). Our method can be seen as a novel reparameterization of a deep model without changing its architecture. This enables us to study the effect of increasing the size of the architecture (both depth and width) without changing the number of optimized parameters (see Figure \ref{num_of_alpha}).

In addition to efficiency, our proposed method provides secure communication and storage, which is of significant interest in applications concerning cybersecurity and privacy. Briefly, our `basis' models are generated with pseudo-random generators with a specific `seed.' This seed could be privately shared between authenticated entities. Given the minimal self-correlation of pseudo-random sequences, a slight seed change produces a drastically different set of `basis' models, making the publicly shared linear mixture coefficients useless to unauthorized parties. This design choice facilitates secure communication and storage, especially in cybersecurity or privacy-sensitive applications.




Theoretically, overparametrization is vital in contemporary neural networks, enhancing their representational power and simplifying optimization due to an improved loss landscape \cite{neyshabur2018the,liu2022loss}. Solutions of these over-parameterized systems often form a positive-dimension manifold \cite{cooper2021global}, with larger systems lacking non-global minima \cite{nguyen2018on}. Considering the abundance of good solutions, we examine if we can confine the solution search to low-dimensional subspaces defined by random vectors in the weight space (i.e., the `basis' networks). Our experiments confirm the possibility of finding good solutions in very low-dimensional random subspaces in the weight space of overparametrized networks, urging further theoretical investigations.


\vspace{.1in}
\noindent{\bf Contributions:} Below are our specific contributions: 
\vspace{-.1in}
\begin{itemize}
    \setlength\itemsep{-0.5em}
    \item Introducing PRANC, a simple but effective network reparameterization framework that is memory-efficient during both the learning and reconstruction phases,
    \item Assessing the effectiveness of PRANC in compacting image recognition models for various benchmark datasets and model architectures, showing higher accuracy with a much fewer parameters compared to extensive recent baselines,
    \item Demonstrating the effectiveness of PRANC for image compression by compacting implicit neural representations for both natural and medical images, 
    \item Showcasing the potential of PRANC in applications requiring encrypted communication of models (or data represented via models).
\end{itemize}


\section{Related work}
{\bf Random networks:} Some prior works \cite{hidden_random, malach2020proving,chen2021peek,gallicchio2020deep} have shown that randomly initialized networks have a subnetwork that competes with the original network in accuracy. Some recent papers like \cite{wortsman2020supermasks} introduced an application for using this fact in continual learning. 
Instead of finding subnetworks in a randomly generated network (i.e., masking), we seek a linear combination of a small set of randomly generated networks, denoted as \emph{basis} models, that can solve the task.   

\textbf{Model compression:}
Model compression is not a new topic. HashedNet \cite{chen2015compressing} uses weight grouping with a hash function to reduce the number of learnable parameters. It can be seen as a specific case of our method where the random models are binary with an equal number of ones and each weight of the original model is one only in one of the random models. \cite{chen2015compressing} experiments with MLP on small datasets. We reproduce HashedNet for our setting and show that our method outperforms it.
Similar to HashedNet, Weight Fixed Network (WFN) \cite{subia2022weight} compresses the model by minimizing the entropy and number of unique parameters in a network. WFN preserves the model's accuracy with a $10\times$ reduction in storage size. Instead of hard-sharing the weights in HasedNet, \cite{ullrich2017soft} uses soft sharing. Although all these methods reduce the number of parameters, they all need to keep the index of each element to reconstruct the network. 
Han \etal \cite{han2015deep} use pruning, quantization, and Huffman coding to achieve compression rates generally less than $50\times$. 
More recent approaches like MIRACLE \cite{havasi2018minimal} and weightless \cite{reagan2018weightless} have shown promising results with much higher compression rates (\eg, $+400\times$). However, they use large architectures, \eg VGG which has 150M parameters, so even after $400\times$ compression, there are still more than 300K parameters (more than a dense ResNet-20).
We show that we can reduce the number of required parameters keeping the network architecture intact.

\textbf{Model pruning and quantization:}
Compressing a model can be defined as reducing the number of bytes required to store a deep model. Several papers like XNOR-NET \cite{rastegari2016xnor} and EWGS \cite{lee2021network} use weight/activation (W/A) quantization for reducing the size of a network. Although W/A Quantization has proven to be an effective approach for reducing network size while maintaining accuracy, it is mainly designed for optimizing the computation for network inference. 
Another approach that is used for compressing a model is pruning the set of less important weights to zero, which reduces the number of floating point operations (FLOPS) and can also reduce the amount of data required to store and communicate a network. These methods include: Neuron Merging \cite{kim2020neuron}, Dynamic pruning \cite{lin2020dynamic,siems2021dynamic}, ChipNet \cite{tiwari2021chipnet}, Pruning at initializing \cite{hayou2020robust}, Wang et.al. \cite{wang2020neural}, and Collaborative Compression (CC) \cite{li2021towards}. Once again, most of these methods use sparsity factors of $20\times$ or less, which is lower than our goal in this paper. We compare our method, PRANC, with existing works that provide extreme compression rates (+99\% pruning rate), e.g., DPF\cite{lin2020dynamic}, STR\cite{str}, and SuRP\cite{isik2022information}. 
Lastly, there are some prior works that decompose model filters as a linear combination of some basis filters \cite{han2020ghostnet,bagherinezhad2017lcnn}. The goal of such methods is to reduce the computation and not necessarily the number of parameters. We focus on an extremely small number of parameters that cannot be achieved by such methods.

\textbf{Data compression - core set:}
Another approach to recreating an accurate network is to store or communicate its training dataset and train a network in the target agent. Since most of the datasets are large, methods are proposed to synthesize metadata in the shape of images or obtain a core set of the dataset. These methods include: Dataset Distillation (DD) \cite{dataset_distil}, which regresses images and learning rate, Flexible Dataset Distillation (FDD) \cite{fdd}, which regresses pseudo-labels for real images, soft labeling dataset distillation (SLDD) \cite{sldd}, that generates pseudo-label and images. All these methods require the seed that initializes the network. Other methods, including Dataset Condensation with distribution matching (DM)  \cite{dm}, with differentiable Siamese augmentation (DSA) \cite{dsa}, and Dataset distillation by matching training trajectories (DDMT) \cite{ddmt} took a step further and devised seed-independent approaches. These methods often rely on a second-order optimization, which is computationally expensive and limits their application.
Moreover, the size of data required for storage in these methods is proportional to the size of input images. We show that PRANC provides better accuracy with a much fewer regressed parameters on the same architectures compared to the mentioned approaches. 

\textbf{Image compression:}
Some popular codecs like JPEG are based on hand-crafted modules to compress an image. Another line of image compression methods is learning-based approaches. These approaches usually train an auto-encoder on a large population of images \cite{balle2018variational,minnen2018joint,lee2018context,djelouah2019content,guo2020variable} and store the code. 
Our method of using INR is also learning-based but is different from the above techniques since the model is learned on a single image (to overfit) rather than on a population of images. Hence, it may not suffer from the biases of the training data. COIN \cite{dupont2021coin} is probably the closest to our method, which overfits an INR and stores all the parameters. We are different since we compact the INR by reparametrizing it as a linear combination of random networks and storing the coefficients. 


\section{Method}
We are interested in training a deep model with a very small number of parameters so that it is less expensive to transfer the model from one agent to another or store it on an agent with small memory. This is in contrast to the goal of most prior works (\eg, model compression, pruning, or quantization) that aim to reduce the inference computation or improve the generalization while preserving the accuracy. Hence, we introduce a compact representation assuming no change in the model size, number of non-zero parameters, or the precision of computation.

We assume that the deep model can be written as a linear combination of a set of randomly initialized models, called {\bf basis}. Since we can use a pseudo-random generator to generate the random models, one can communicate or store all random models by simply communicating or storing a single scalar that is the seed of the pseudo-random generator. Although basis models are not necessarily orthogonal to each other, their pairwise dot product is close to zero since the number of samples (models) is much smaller than the dimensionality of each model. Then we optimize the weights of each base model so that their linear combination can solve the task (e.g., image classification).

\begin{figure}[t!]
    \centering
    \includegraphics[width=.98\linewidth]{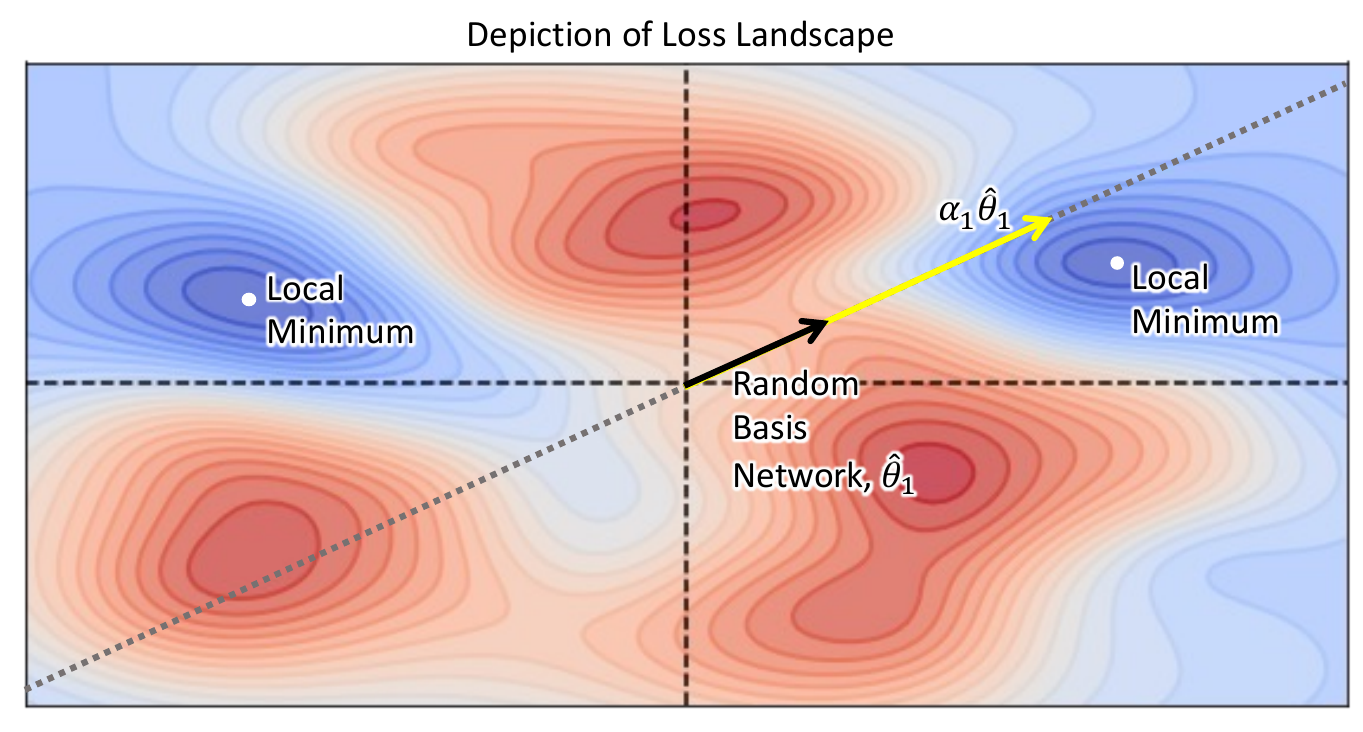}
    \vspace{-.1in}
    \caption{A simple illustration of the loss landscape of a model with two parameters and one basis model. None of the two local minima may be in the span of the basis models, so we search for $\alpha$ to find a local minimum in the span of the basis models.}
    \label{fig:my_label}
    \vspace{-.2in}
\end{figure}

More formally, given a set of training images $\{x_i\}_{i=1}^N$ and their corresponding labels $\{y_i\}_{i=1}^N$, we want to train a deep model $f(.;\theta)$ with parameters $\theta \in \mathbb{R}^d$ so that $f(x_i;\theta)$ predicts $y_i$. The standard practice is to optimize $\theta$ by minimizing the empirical risk: 
\begin{center}
    $R(\theta)=\frac{1}{N}\sum_{i=1}^N L(f(x_i;\theta), y_i)$
\end{center}

where $L(\cdot,\cdot)$ is a discrepancy-measure, e.g., cross-entropy. In communicating such a model, we need to send a high-dimensional vector $\theta$ that contains $d$ scalars.

To reduce the cost of communication, we assume a set of randomly initialized basis models with parameters $\{\hat{\theta}_j\}_{j=1}^k$. These $k$ basis models are generated using a known seed and are frozen throughout the learning process. Then we define: $\theta := \sum_{j=1}^k \alpha_j \hat{\theta}_j$,  where $\alpha_j$ is a scalar weight for the $j$'th basis model. Assuming that $k \ll d$, it will be much less expensive to communicate or store $\alpha$ instead of $\theta$.

To optimize $\alpha$, one may first optimize for $\theta$ to find $\theta^*$ and then regress it by minimizing:
\begin{center}
    $\argmin_{\alpha} ||\theta^* - \sum_{j=1}^k \alpha_j \hat{\theta}_j||^2$ 
\end{center}

However, since $k \ll d$, the optimum solution $\theta^*$ may be far from the span of the basis models, resulting in an inferior solution (also shown empirically in our experiments). We argue that there are an infinite number of solutions for $\theta$ that are as good as $\theta^*$, so we may search for one with a smaller residual error when projected to the span of the basis models. Hence, we search for a solution that minimizes the task loss in the basis models' span by optimizing:
\begin{align}
    \argmin_\alpha \sum_i L\Big(f(x_i;\sum_{j=1}^k \alpha_j \hat{\theta}_j), y_i\Big)
\end{align}


Note that at the test time, after reconstructing the model by linear combination, the inference for PRANC is exactly the same as the standard dense model.

{\bf Optimization efficiency:} Note that the optimization is very simple and efficient since $\frac{dL}{d\alpha} = \frac{dL}{d\theta} \frac{d\theta}{d\alpha}$ and $\frac{d\theta}{d\alpha_j} = \hat{\theta}_j$. Hence, we use standard backpropagation to calculate $\frac{dL}{d\theta}$ and then multiply that with the basis models' matrix to get:
$$\frac{dL}{d\alpha} = \frac{dL}{d\theta} \times \hat{\theta}$$

{\bf Memory efficiency in training:}
Note that the matrix of basis models $\hat{\theta}$ is very large, so keeping that in the memory is not efficient. Hence, we divide this matrix into multiple smaller chunks, and at each iteration, we generate each chunk using a pseudo-random generator at the GPU itself, perform the multiplication, discard the chunk, and go to the next chunk. This method reduces the memory footprint by a large factor at the cost of generating the whole random basis once per iteration, which is very efficient in modern GPUs. Choosing chunks of 100 alpha values for ResNet18 consumes almost 4.4GB (i.e., $11M\times 4 \times 100$) of GPU memory which is reasonable.

{\bf Model reconstruction efficiency:} Since basis models are generated using a pseudo-random generator, we can reconstruct the model using a simple running average of the basis models: generate each entry in $\hat \theta_j$, multiply it with $\alpha_j$, add it to the running average, discard the entry and go to the next entry of $\hat \theta_j$. This way, the memory footprint of the reconstruction becomes negligible (i.e., $d+1$). 

{\bf On-demand model reconstruction:} In some applications, the agent may need to run the inference rarely but does not have enough memory to hold the model. The device can store $\alpha$, reconstruct each convolutional filter using the corresponding entries of the basis models, apply it to the input, and then discard the filter and go to the next filter. This process has a very small memory footprint as it needs to store $\alpha$ and just one filter at a time.

{\bf Distributed learning:} In order to train the model on multiple GPUs, we use a simple distributed learning algorithm to increase $m$, the number of basis models. We divide $m$ basis models between $g$ GPUs so that each GPU works on $m/g$ basis models only. Then, we distribute $\alpha$ among GPUs. Each GPU calculates the partial weighted average over its basis models and distributes it to all GPUs. Then, all GPUs will have access to the complete weighted average and will use it to do backpropagation in standard distributed learning form and update their own set of $\alpha$.

{\bf BatchNorm layer:} We minimize the loss of the task by tuning the $\alpha$ instead of the model weights as done in standard learning. However, the parameters of the BatchNorm layer are not tuned by PRANC. For the simplicity of this work, we assume that we can communicate those parameters and include them in the budget. This makes sense since the number of BatchNorm parameters is relatively small compared to the number of weight parameters. 

\section{Application}
We test our framework on two different applications.

{\bf Image classification networks:}
In this setting, we parameterize an image recognition model, e.g., ResNet-20 for CIFAR-10, by our PRANC framework and optimize $\alpha$s instead of the model weights. This results in a compact model that can be stored and communicated very efficiently.  

{\bf Image compression using implicit neural networks:}
We also test our framework on compressing an implicit neural network (INR) that is over-fitted to a single image \cite{tancik2020fourier}. Such an INR inputs the coordinate of the pixel and returns the color value. Hence, one can store or communicate the INR model instead of the original image. We parameterize a standard INR model \cite{tancik2020fourier} using the PRANC framework so that we optimize the $\alpha$s instead of the weights of the INR model. Our method outperforms JPEG compression on two standard datasets and two evaluation metrics. 

\section{Experiments on image classification}
We report extensive results of PRANC on various datasets, architectures, and number of basis models. 



\subsection{Comparison with model pruning methods:}
For communicating a sparse model with a sparsity rate of more than $2\times$, it is required to transmit two numbers per parameter: the value of weight and its index in the network. Therefore, even if a model pruning method uses a pruning factor of 99\% ($100\times$ reduction in size) since it should transmit the indices alongside the values, the actual reduction size will be smaller than $100\times$.
DPF \cite{lin2020dynamic}, STR\cite{str}, LAMP\cite{lee2020layer}, RiGL\cite{evci2020rigging}, and SuRP\cite{isik2022information} are the SOTA methods that use a large sparsity $(+50\times)$ and maintain a reasonable accuracy. We used their code on CIFAR-10 and CIFAR-100 along with ResNet-20 and ResNet-56 architectures and compared them with our method in Table \ref{tab:compress}. PRANC achieves consistently higher accuracy with fewer parameters. Please note that all these methods excluded BatchNorm layers from their pruning process. Therefore we also excluded them from the parameter count. For ResNet-20, the number of BatchNorm parameters is 2,752 and for ResNet-56 it is 8,128.

\begin{table}[h]
  \caption{Comparison of our model with SOTA pruning methods, DPF \cite{lin2020dynamic}, STR\cite{str}, LAMP\cite{lee2020layer}, RiGL\cite{evci2020rigging}, and SuRP\cite{isik2022information}. ``Pr.'' denotes the pruning rate. Also, when the network is pruned, we have to keep two numbers for each weight: the weight itself and its position in the model. Note that we excluded the number of BatchNorm parameters in this table since that is constant for all the models. This number is 2,752 for Resnet-20 and 8,128 for ResNet-56.}
  \centering
   \label{tab:compress}
   {\footnotesize
  \begin{tabular}{|c|c|c|c|c|}
    \hline & &&& \\[-1.5ex]
    \textbf{Method} & \textbf{Data} & \textbf{Arch.}  & \textbf{\# Params exc.} & \textbf{Accuracy}   \\
    & && \textbf{BatchNorm} &    \\
    \hline & &&& \\[-1.5ex]
    Baseline (Pr. 0\%) & C10& R20& 269,722 & 88.92\\
    DPF(Pr. 98.2\%) & C10 & R20 & 4,920$\times$2 & 41.86 \\
    RiGL(Pr. 99.62\%) & C10 & R20 & 1026$\times$2 & 50.9\\
    LAMP(Pr. 99.62\%) & C10 & R20 & 1026$\times$2 & 51.24\\
    SuRP (Pr. 99.62\%) & C10 & R20 & 1026$\times$2 & 54.22\\
    STR (Pr. 95.5\%) & C10 & R20 & 12,238$\times$2 & 75.99 \\
    \textbf{Ours} & \textbf{C10} & \textbf{R20} & \textbf{1,000} & \textbf{64.59 } \\
    \textbf{Ours} & \textbf{C10} & \textbf{R20} & \textbf{10,000} & \bf{81.48} \\
    \hdashline & &&& \\[-1.5ex]
    Baseline (Pr. 0\%) & C10& R56& 853,018 & 91.64\\
    DPF (Pr. 98.43\%) & C10 & R56 & 13,414$\times$2 & 47.66 \\
    SuRP (Pr. 98.73\%) & C10 & R56 & 10,834$\times$2 & 66.65 \\
    STR (Pr. 98.4\%) & C10 & R56 & 13,312$\times$2 & 67.77 \\
    \textbf{Ours} & \textbf{C10} & \textbf{R56} & \textbf{5,000 } & \textbf{76.87}\\
   \hline & &&& \\[-1.5ex]
    Baseline (Pr. 0\%) & C100& R20& 275,572 & 60.84\\
    DPF (Pr. 96.13\%)& C100 & R20 & 10,770$\times$2 & 12.25\\
    SuRP (Pr. 97.48\%) & C100 & R20 & 6,797$\times$2 & 14.46 \\
    STR (Pr. 96.12\%)& C100 & R20 & 10,673$\times$2 & 13.18 \\
    \textbf{Ours} & \textbf{C100} & \textbf{R20} & \textbf{5,000 } & \textbf{32.33}  \\
    \hdashline & &&& \\[-1.5ex]
    Baseline (Pr. 0\%) & C100& R56& 858,868 & 64.32\\
    DPF (Pr. 97.8\%) & C100 & R56 & 19,264$\times$2 & 19.11\\
    SuRP (Pr. 98.72\%) & C100 & R56 & 10,919$\times$2 & 14.59 \\
    STR (Pr. 97.8\%) & C100 & R56 & 18,881$\times$2 & 25.98 \\
    \textbf{Ours} & \textbf{C100} & \textbf{R56} & \textbf{5,000 } & \textbf{32.97}\\
    \hline
  \end{tabular}}
  \vspace{-.2in}
\end{table}


\subsection{Comparison with model distillation methods:}
One of the critical baselines for our work is model distillation. However, the number of parameters we use is very small compared to any existing CNN architecture. Even LeNet\cite{lecun2015lenet} (one of the smallest CNN architectures), has more than 60,000 parameters. To compare PRANC with model distillation, we trained a ResNet18 on CIFAR-10 and distilled its knowledge to a LeNet model. On the other hand, we compressed a ResNet20 model using PRANC with 10,000 $\alpha$s and a ResNet56 model with merely 5,000 $\alpha$s and compared their accuracies. As shown in Table \ref{tab:distil}, PRANC-compressed architectures require almost $5\times$ fewer parameters while achieving higher accuracies with a significant gap (81.48\% vs. 74.1\%).
\begin{table}[h]
  \caption{Comparison with model distillation. PRANC outperforms a LeNet distilled from ResNet-18 on CIFAR-10. 2,752 and 8,128 are the number of BatchNorm parameters that we exclude from the coefficients but need to consider them as parameters.}
  \centering
   \label{tab:distil}
   {\small
  \begin{tabular}{|c|c|c|c|}
    \hline 
    \textbf{Method}  & \textbf{Arch.}  & \textbf{\# Params} & \textbf{Acc.}   \\
    \hline & && \\[-1.5ex]
    Distilled from R18 & LeNet & 62,006 & 74.1\%\\
     \textbf{Ours} & \textbf{R56} & \textbf{5,000 + (8,128)} & \textbf{76.87\%}\\
    \textbf{Ours} & \textbf{R20} & \textbf{10,000 + (2,752)} & \textbf{81.48\%}\\
    \hline
  \end{tabular}}

\end{table}



\subsection{Comparison with dataset distillation methods:}
In Table \ref{tab:data_distil}, we report the accuracy and the number of parameters for PRANC in comparison with various dataset distillation methods. Most of these methods are based on meta-learning approaches that involve a high computational cost and memory footprint at the training time, so they are limited in the depth of the model. Moreover, they need to do a few gradient descent steps in constructing the model. Nonetheless, the number of required parameters in dataset distillation methods is proportional to the size of the input image. For instance, in distilling CIFAR-10 to 10 images only, we need to store at least $10 \times 32 \times 32 \times 3$ parameters. In order to be comparable with the SOTA Dataset Distillation methods,  we use AlexNet (which is a modified version that is described in \cite{dataset_distil}) on CIFAR-10. For CIFAR-100 and tinyImageNet, we use depth-3 and depth-4 128-width ConvNet architectures described in \cite{ddmt}, respectively. Note that some dataset distillation methods do not require a seed, so they solve a more challenging task since the distilled data should be able to tune any randomly initialized model. However, since we are focusing on reducing the cost of communication and storage, using a fixed seed as the central part of our idea is not prohibitive.



\begin{table}[h]
  \caption{Comparison with dataset distillation methods on various datasets and architectures. 3-128-Conv and 4-128-Conv represents 3-depth 128-width ConvNet and 4-depth 128-width ConvNet, respectively. ``Trained model" is the upper bound of our method since one can optimize all weights and transmit/store them. Our method outperforms the baselines with a large margin and a much fewer parameters. }
  \centering
   \label{tab:data_distil}
   \scalebox{0.87}{
  \begin{tabular}{|c|c|c|c|c|}
    \hline & &&& \\[-1.5ex]
    \textbf{Method} & \textbf{Task} & \textbf{Arch}  & \textbf{\# Params} & \textbf{Acc.} \\
    \hline & &&& \\[-1.5ex]
    Trained model & C10 & AlexNet & 1,756,426 & \textbf{$84.8$} \\
\hdashline   & &&& \\[-1.5ex]
    FDD \cite{fdd} & C10 & AlexNet & 397,000 & $43.2 $ \\
    SLDD \cite{sldd}& C10 & AlexNet& 308,200 & $60.0$ \\
    DD \cite{dataset_distil} & C10 & AlexNet& 307,200 & $54.0 $ \\
    DM \cite{dm} & C10 & AlexNet& 30,720 & $26.0$ \\
    DSA \cite{dsa} & C10 & AlexNet& 30,720 & $28.8 $ \\
    DC \cite{dc} & C10  & AlexNet& 30,720 &  $28.3$ \\
    CAFE \cite{cafe}& C10 & AlexNet& 30,720 &  $30.3 $ \\
    CAFE+DSA \cite{cafe}& C10 & AlexNet & 30,720 & $31.6$ \\
    DDMT \cite{ddmt} & C10 & AlexNet & 30,720 & $46.3 $ \\
     \textbf{Ours} & C10 & AlexNet & \textbf{17,000} & \textbf{76.69} \\
    \hline & &&& \\[-1.5ex]
    Trained model & C100& 3-128-Conv & 504,420 & \textbf{$56.2 $} \\
    \hdashline &&&& \\[-1.5ex]
    FDD \cite{fdd} & C100 & 3-128-Conv & 317,200 & $11.5 $ \\
    DM \cite{dm}& C100 &  3-128-Conv & 307,200 & $11.4$ \\
    DSA \cite{dsa} & C100 &  3-128-Conv & 307,200 & $13.9 $ \\
    DC \cite{dc} & C100 & 3-128-Conv & 307,200 & $12.8 $ \\
    CAFE+DSA \cite{cafe}& C100  & 3-128-Conv & 307,200 &  $14.0 $ \\
    DDMT \cite{ddmt} & C100 & 3-128-Conv & 307,200 & $24.3$ \\ 
    
    \textbf{Ours} & C100 & 3-128-Conv& \textbf{15,000} & \textbf{25.57} \\
    \hline & &&& \\[-1.5ex]
    Trained model & tinyIN& 4-128-Conv & 857,160 & \textbf{$37.6 $} \\
    \hdashline &&&& \\[-1.5ex]
    DM \cite{dm}& tinyIN &4-128-Conv& 2,457,600 & $3.9$ \\
    DDMT \cite{ddmt} & tinyIN  &4-128-Conv& 2,457,600 &  $8.8 $ \\
    \textbf{Ours} & tinyIN& 4-128-Conv & \textbf{15,000} & \textbf{12.02} \\
    \hline
  \end{tabular}
  }
  \vspace{-.1in}
\end{table}

\subsection{Large-Scale dataset and models:}
So far, we have provided evidence that our method can outperform recent works in dataset distillation, model pruning, and model compression in terms of the number of parameters vs. accuracy. Since our method is reasonably efficient in learning, particularly compared to meta-learning approaches that depend on second-order derivatives of the network, we can evaluate it on larger-scale models. We evaluate our method on ImageNet100 with ResNet-18 architecture. Table \ref{tab:IN} shows the results. Our method achieves 61.08\% accuracy with less than 1\% parameters, while the standard ResNet-18 model achieves 82.1\% accuracy with more than 11M parameters. Since ResNet-18 is a huge model, we use one of the unique capabilities of PRANC, which is creating the network on the fly. We only used a single NVIDIA 3090 GPU and trained our model for 200 epochs, using Adam optimizer and step scheduler with $\gamma = 0.5$ for every 50 epochs and an initial learning rate of 0.001. Also, we distributed our budget of $\alpha$ vector across the layers, \textit{i.e.,} we used 4,000 coefficients for each layer of the convolutional encoder and 20,000 coefficients for our classifier (giving us a total of 100,000 coefficients).

\begin{table}[h]
  \caption{Result of our method on ImageNet-100 dataset and ResNet-18. 19,200 is the total number of parameters of all BatchNorm layers in our model}
  \centering
   \label{tab:IN}
   \small
  \begin{tabular}{|c|c|c|}
    \hline & & \\[-1.5ex]
    \textbf{Method}   & \textbf{\# Params} & \textbf{Acc.}   \\
    \hline & & \\[-1.5ex]
     trained model  & 11,227,812 & \textbf{82.1\%} \\
     \hdashline   & & \\[-1.5ex]
     HashedNet \cite{chen2015compressing}  & 110,000 + (19,200) & 52.96\% \\
    \textbf{Ours} & \textbf{100,000 + (19,200)} & \textbf{61.08\% }  \\
    \textbf{Ours} & \textbf{200,000 + (19,200)} & \textbf{67.28\% }  \\
    \hline
  \end{tabular}
\vspace{-.1in}  
\end{table}



\subsection{Ablation study:}
In PRANC, $k$, the number of basis models, is a hyperparameter. One question is: ``How will $k$ affect the performance?'' Moreover, it is arguable that ``why do we try to find a linear combination that is accurate on the task? why not try to regress an entire already trained network?'' Also, ``Can $k$ be more important than the architecture? \textit{i.e.,} can we use large $k$ with a small network and still get high accuracy?" In this section, we answer these questions.

{\bf Sensitivity to seed:} Since one of the applications of PRANC is in federated learning, it is worth discussing its pseudo-encryption ability. We experimented with changing the seed at the reconstruction time. On CIFAR-10 with AlexNet, with a minor change in seed, the accuracy of the reconstructed model dropped from \textbf{74.0\%} to \textbf{9.4\%}, which is close to chance. This is expected as the new basis models are not correlated with the ones that are used in training. This fact means that the seed can act as a mutual key between the agents and even if $\alpha$ is intercepted, reconstructing the model is nearly impossible. Therefore, in federated learning applications which deal with safe communication of deep learning models, PRANC can be used as both \textbf{compression} and \textbf{encryption} method.



{\bf Regressing $\theta^*$ directly:}
We can first train a model to get $\theta^*$ and then optimize for $\alpha$ by regressing that solution using MSE loss in the parameter space. As shown in Fig. \ref{fig:my_label}, this may not succeed since the optimal model may not be in the span of the basis models, and also the MSE loss in the parameter space is not necessarily correlated with the task loss. Table \ref{theta_star} shows that the accuracy of this baseline using 10,000 parameters is close to chance.

\begin{table}[h]

  \caption{Results of regressing a pretrained model using 10,000 basis models. C10 and C100 denote CIFAR-10 and CIFAR-100 and tinyIN denotes tiny ImageNet datasets }
  \centering
     \label{theta_star}
  \begin{tabular}{|c|c|c|c|}
    \hline 
    \textbf{Dataset} & \textbf{Architecture} & \textbf{Full Acc.} & \textbf{Regress. Acc.} \\
    \hline 
    C10 & AlexNet &  84.8 \% & 10.0\% \\
    C10 & LeNet \cite{lecun2015lenet} & 73.5\% & 12.74\% \\
    \hdashline 
    C100 & 3-128-Conv & 56.2 \% & 1.14\% \\
    C100 & AlexNet & 50.7\% & 1.0\% \\
    \hdashline 
    tinyIN & 4-128-Conv &  37.6 \%  & 0.5\% \\
    \hline
  \end{tabular}

\end{table}

\begin{figure}[h]
  \centering
  \includegraphics[width=0.48\textwidth]{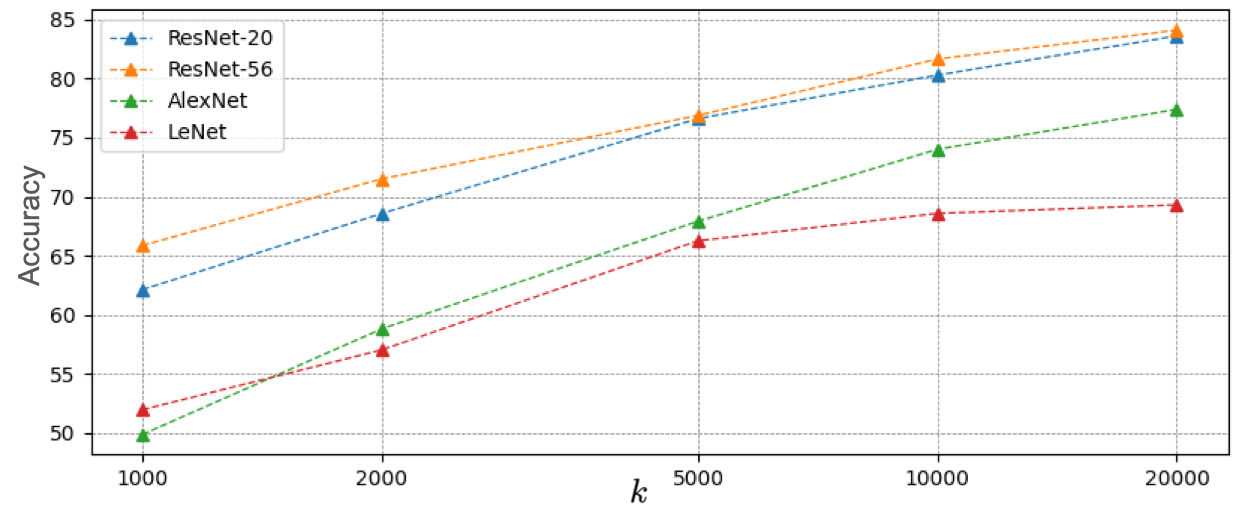}
  \caption{Illustration of impact of $k$ in accuracy of different architectures on CIFAR-10. The accuracy improves by increasing the number of basis models. However, the architecture plays more critical role in the accuracy compared to the number of basis.}
  \label{num_of_alpha}
\end{figure}

{\bf Effect of varying $k$ vs. architecture:} 
We perform an ablation study to understand the effect of the number of basis models, $k$ vs. the architecture of the model. One can argue that sometimes it is better to design a better architecture rather than increasing the number of $\alpha$s. Here, we change $k$ from 1,000 to 20,000 for LeNet, AlexNet, ResNet-20, and ResNe-56 on CIFAR-10 and plot the accuracy in Figure \ref{num_of_alpha}. All experiments have been done in the same setup with 400 epochs with Adam optimizer. As expected, the accuracy increases as we increase the number of basis functions. However, as we can see, the architecture has more effect on the accuracy compared to $k$. 

\section{Experiments on image compression}
As another application of PRANC, here we show that we can compress an image by compacting the implicit neural network (INR) that is overfitted to the image. The INR model inputs the coordinates of a pixel and outputs the color of that pixel. We will store or communicate the seed of the pseudo-random generator and $\alpha$ values only. We use \cite{tancik2020fourier} as our INR model. The INR is a 4-layer MLP (3 hidden layers) model with 512 neurons at each layer and a Layer-Norm \cite{ba2016layer} after each layer. We encode the pixel coordinate in the input to 512 dimensions using Fourier mapping and use three neurons in the output for RGB images (the color values) and one neuron for X-ray images. 

We use a half-precision floating point for $\alpha$s to further reduce the storage cost. We use a different set of $\alpha$s for each layer of the network. 

To reduce the memory footprint, we divide the base network matrix $\hat{\theta}$ into smaller chunks and generate and then discard each chunk at the GPU for each iteration of the optimization. 
Similar to stochastic gradient descent, we randomly sample a subset of pixels to be optimized at each iteration, leading to faster convergence due to more frequent parameter updates.

We evaluate our model on three different datasets: {\bf Kodak} dataset \cite{kodakdataset} that contains 24 non-compressed images of size $512 \times 768$, {\bf CLIC-mobile} test set \cite{CLIC2020} that contains $178$ high-resolution images (e.g., $1512 \times 2016$), and  $64$ randomly sampled images from {\bf NIH Chest X-ray} dataset \cite{wang2017chestx} that consists of $100,000$ de-identified chest X-rays images of size $1024 \times 1024$. 



\textbf{Baselines.} We compare our method with the following hand-crafted image codecs: JEPG, JPEG2000, and WebP. Our goal is to show that PRANC is a general framework that performs well when simply applied to INR compression out of the box. Hence, we do not compare it with more advanced codecs like BPG and VTM since they are highly engineered and include components like entropy coding. Those results are presented in the supplementary material. We also compare with learning-based image compression methods in the supplementary (BMS, MBT, and CST). Note that, unlike PRANC, these methods require a large dataset of images to pre-train their auto-encoder. This limits the applications and also may degrade the results for out-of-distribution data points. For instance, unlike PRANC, models that are trained on a training set may not be suitable for medical data since they may not be truthful to abnormalities specific to patients not represented in the training data.


COIN \cite{dupont2021coin} is probably the closest to our method which trains an INR using SIREN \cite{sitzmann2020implicit} and stores all the parameters. Since COIN does not use Fourier mapping, for a fair comparison, we produce a similar baseline, called `trained INR,' using our MLP network described above without the PRANC framework. For the trained INR, we reduce the width of the network to match the compression rate of our method and use half-precision floating points.


\textbf{Results.} We evaluate our model with two metrics: PSNR and MS-SSIM. Note that we fix the network architecture and vary the number of $\alpha$ values to get different bit-per-pixel (bpp) values for each image. We report the results for the Kodak dataset in Figure \ref{fig:kodak} and CLIC-mobile dataset on Table \ref{tab:clic_mobile}. Our method outperforms JPEG and INR. We report the results of the chest X-ray in Table \ref{tab:xray}. An example image is shown in Figure \ref{fig:kodak_vis} for the Kodac dataset and in Figure \ref{fig:xray_1} for the X-ray dataset.

\textbf{Ablation.} Since we reconstruct the model weights with $\alpha$ values, one can vary the size of the architecture without changing the number of parameters ($\alpha$ values), hence, in PRANC, we can increase both width and depth without changing the bit-rate. We keep the number of parameters $k=102K$ and vary both the network width and depth of the MLP model. Results on the Kodak dataset are shown in Table \ref{tab:ablation_width_alpha}. Interestingly, we can improve the performance by increasing the model depth while keeping the number of learnable parameters constant.

\textbf{Sorting $\alpha$ values.} In Figure \ref{fig:vis_sort_0}, we show reconstructed images with a subset of $\alpha$ values with the largest absolute values. Since we have a different set of $\alpha$s for each layer of MLP in image compression, we sort absolute values and select the top $p\%$ of each layer with higher values. We vary $p$ and visualize the reconstructed images for each $p$ value. In the supplementary material, for the image classification setting, we show that in reconstructing the ResNet model using a partial set of $\alpha$ values, choosing larger $|\alpha|$ values leads to much better accuracy compared to choosing a random subset.

 
\textbf{Implementation Details.} For each image, we train $\alpha$ values with $10k$ iterations on Kodak and $5k$ iterations on CLIC-mobile dataset. Each iteration processes 25\% of the pixels of the image sampled randomly. Note that increasing the number of iterations cannot damage the model since the goal is to overfit to the image and generalization is not an issue. We use PyTorch Adam \cite{kingma2014adam} optimizer with an initial learning rate of $1e-3$ and a Cosine scheduler. More details about the number of $\alpha$ values per layer for each bpp are in the supplementary material. 

\begin{figure}
    \centering
    \includegraphics[width=0.95\linewidth]{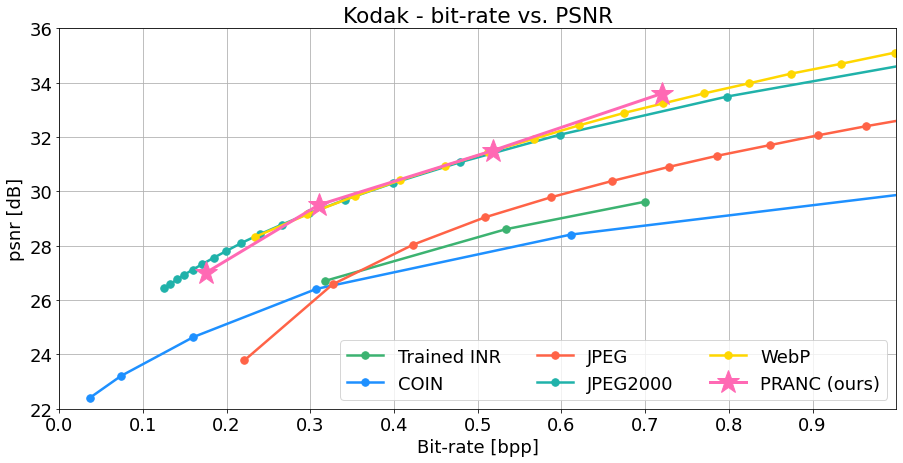}
    \includegraphics[width=0.95\linewidth]{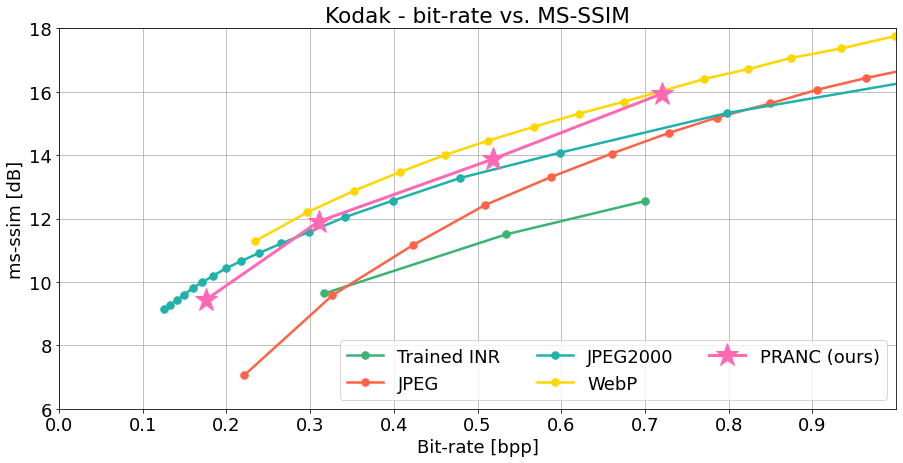}
    
    \caption{\textbf{Kodak Dataset Image Compression:} Our method outperforms JPEG and `trained INR' on both PSNR and MS-SSIM evaluations at various bitrates. Note that, unlike the other baselines, our method is learned on a single image and is not hand-crafted, except for the architecture of the INR model, which is a simple MLP.} 
    \label{fig:kodak}
\end{figure}

\begin{table}
    \begin{center}
    \caption{\textbf{CLIC-mobile Dataset Image Compression:} PRANC outperforms JPEG and JPEG2000 with smaller bpp on this dataset.
    }
    \label{tab:clic_mobile}
    \scalebox{1.0}
    {
    \begin{tabular}{|l|c|c|c|} %
    \hline
    Model & bpp & PSNR & MS-SSIM \\\hline 
    WebP & 0.185 & 30.07 & 0.940 \\
    JPEG2000 & 0.126 & 29.40 & 0.918\\ 
    JPEG & 0.195 & 24.82 & 0.836\\ 
    Trained INR & 0.125 & 26.93 & 0.864 \\
    PRANC (ours) & 0.119 & 29.71 & 0.920\\
    \hline
    
    \end{tabular}
    }
    \end{center}

\end{table}

\begin{table}
    \begin{center}
    \caption{\textbf{Chest X-ray Dataset Image compression:} We compare PRANC with JPEG for X-ray images. Our method is better than JPEG with a lower bpp. Since unlike auto-encoders, PRANC does not use any population-based training, it may be better suited for medical images since it may preserve out-of-distribution artifacts which are important for diagnosis purposes. However, we leave studying this for future work.
    }
    \label{tab:xray}
    \scalebox{1.0}
    {
    \begin{tabular}{|l|c|c|} %
    \hline
    Model & PRANC & JPEG  \\\hline 
    bpp & 0.152 & 0.168 \\
    PSNR & 36.28 & 34.25  \\
    MS-SSIM & 0.972 & 0.921 \\
    \hline
    
    \end{tabular}
    }
    \end{center}
    \vspace{-.1in}
\end{table}

    

\begin{table}
    \begin{center}
    \caption{\textbf{Effect of increasing width/depth of the model:} We can increase both the depth and width of the MLP model without changing the number of parameters. When changing the depth, we redistribute the number of $\alpha$ values uniformly among layers to keep the total number constant. More details are in the Supp. 
    }
    \label{tab:ablation_width_alpha}
    \scalebox{1.0}
    {
    \begin{tabular}{|l|c|c|c|c|} %
    \hline
    Width & 128 & 256 & 512 & 1024\\\hline 
    PSNR & 30.00 & 32.05 & 31.5 & 31.32\\
    MS-SSIM & 0.937 & 0.963 & 0.959 & 0.961\\
    \hline
    
    \end{tabular}
    }
    \scalebox{1.0}
    {
    \begin{tabular}{|l|c|c|c|} %
    \hline
    Depth & 3 & 4 & 5 \\\hline 
    PSNR & 30.78 & 31.5 & 32.38\\
    MS-SSIM & 0.978 & 0.959 & 0.965\\
    \hline
    
    \end{tabular}
    }
    \end{center}
    \vspace{-.1in}
\end{table}

\begin{figure*}[t]
    \centering
    \includegraphics[width=\textwidth]{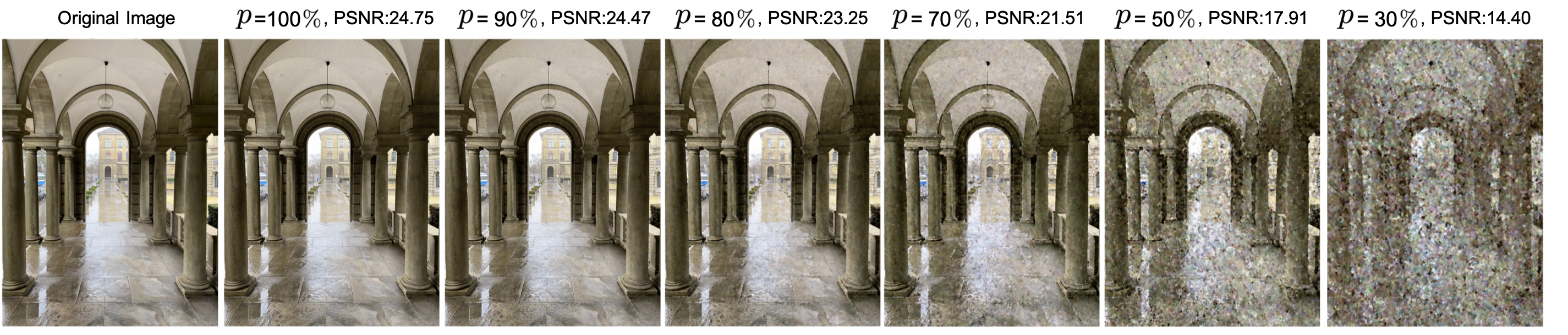}
    
    \caption{\textbf{Effect of keeping $p\%$ of basis models with the highest absolute $\alpha$ values.} We get reasonable images with a smaller subset of basis models. One can reconstruct an approximate image upon receiving a partial set of $\alpha$ values, similar to progressive JPEG.}
    \label{fig:vis_sort_0}
    \vspace{-.15in}
\end{figure*}

\begin{figure}[t]
    \centering
    \includegraphics[width=\linewidth]{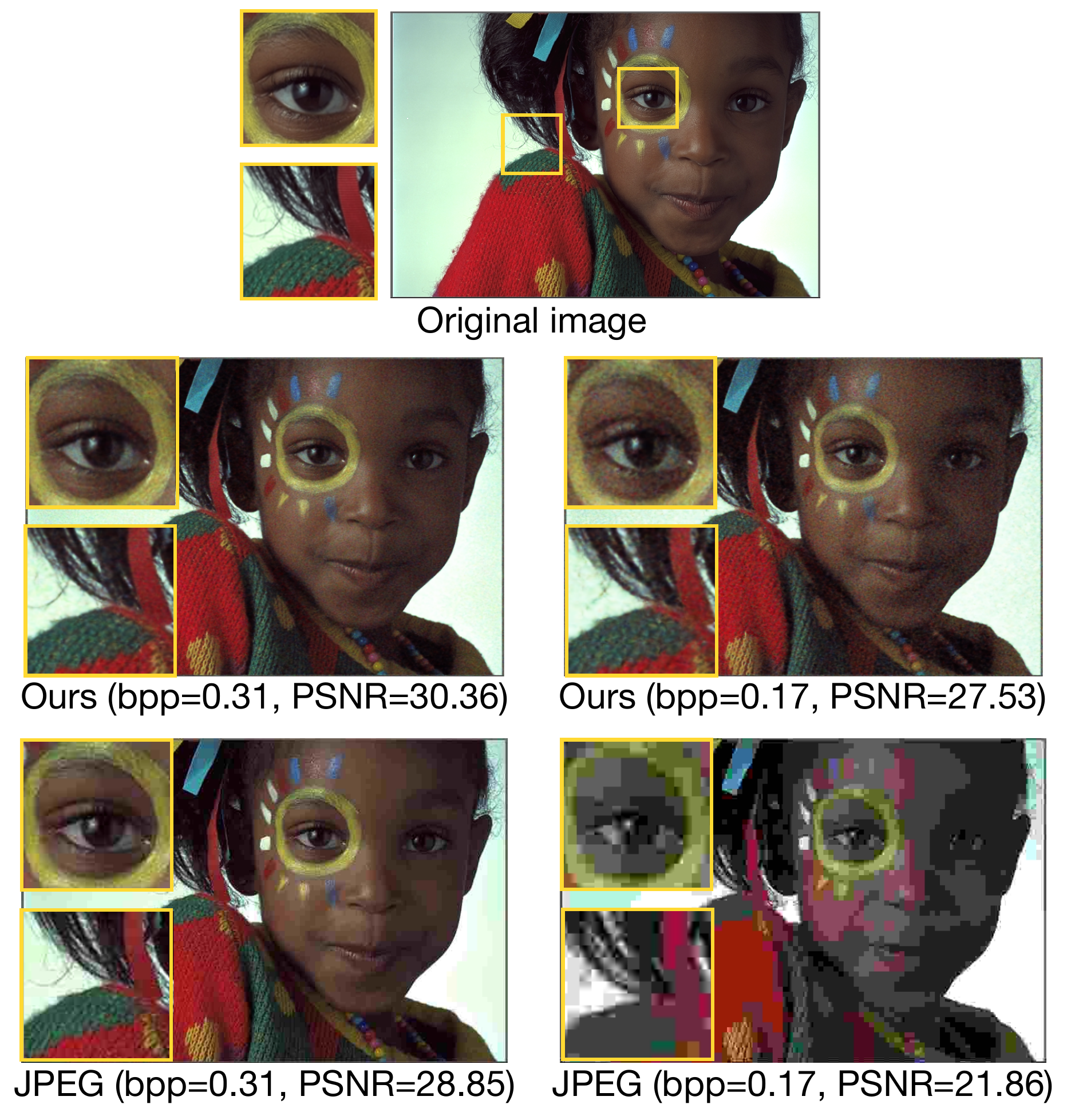}
    \caption{\textbf{Kodak visualization.} We compare PRANC with JPEG on image 15 of the Kodak dataset. See Supp. for more examples.} 
    \label{fig:kodak_vis}
    \vspace{-.15in}
\end{figure}

\begin{figure}[t]
    \centering
    \includegraphics[width=0.95\linewidth]{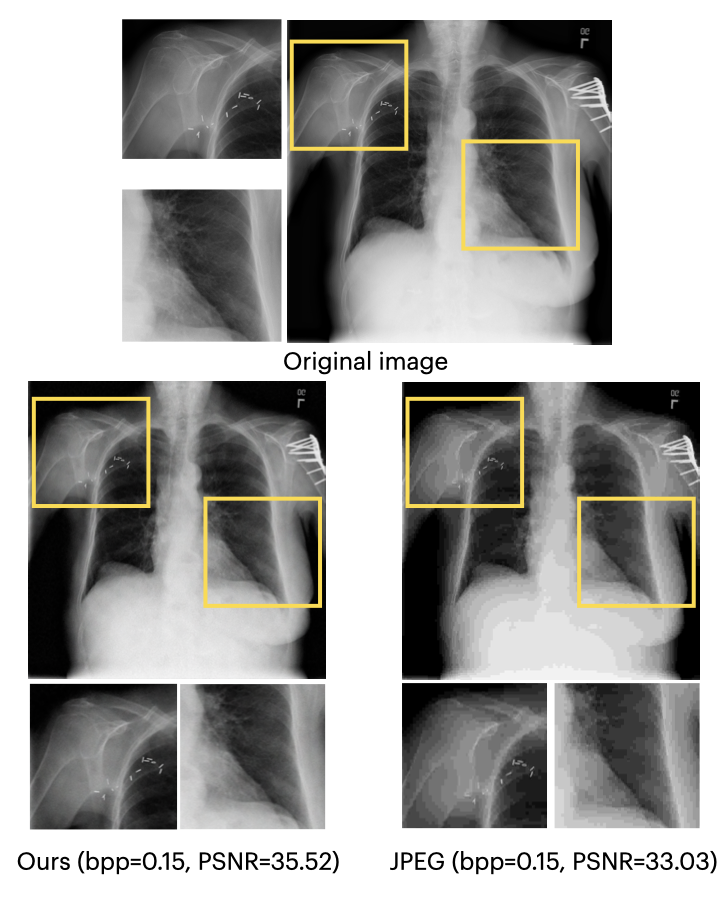}
    \caption{\textbf{Chest X-ray visualization}. We compare PRANC and JPEG on a Chest X-ray image. See Supp. for more examples.}
    \label{fig:xray_1}
    \vspace{-.15in}
\end{figure}


\section{Future directions}
\vspace{-.05in}
\noindent PRANC can enable multiple future directions:
\\
{\bf Generative models for memory-replay:} Our method can be used to compact a generative model (e.g., GANs or diffusion models), where the $\alpha$ parameters may be stored in the agent or sent to another agent. Then, any agent can reconstruct the model in the future and draw samples from it that are similar to the samples that were used earlier to train the model. This enables memory replay in lifelong learning in a single agent with limited memory or in multiple agents with limited communication.

{\bf Progressive compactness:} In this method, we assumed a set of basis models with no specific ordering. However, one can optimize $\alpha$ so that the earlier indices of $\alpha$ can reconstruct an acceptable model. Then, depending on the communication or storage budget, the target agent can decide on how many $\alpha$ parameters it needs by trading off between accuracy and compactness. We showed that sorting $\alpha$ values is a first step in this direction, but one may learn them in decreasing importance order as a future work.


\vspace{-.05in}
\section{Conclusion}
\vspace{-.05in}
We introduced a simple yet effective method that can learn a model as a linear combination of a set of frozen randomly initialized models. The final model can be compactly stored or communicated using the seed of the pseudo-random generator and the coefficients. Moreover, our method has a small memory footprint at the learning or reconstruction stages. We perform extensive experiments on multiple image classification datasets with multiple architectures and also on image compression settings and show that our method achieves better accuracy with fewer parameters compared to SOTA baselines. We believe many applications including lifelong learning and distributed learning can benefit from our ideas. Hence, we hope this paper opens the door to studying more advanced compacting methods based on frozen random networks.

{\bf Limitations:}
As discussed, some model parameters, e.g., BatchNorm layers, cannot be easily reparameterized using our method since they are calculated directly from data rather than minimizing the task loss. In this paper, we assumed we communicate them with no change and included them in our budget. Lastly, PRANC is computationally expensive for a large number of basis, so is currently not suitable for training very large models.

{\bf Acknowledgement:}
This work was partially supported by the Defense Advanced
Research Projects Agency (DARPA) under Contract No. HR00112190135 and HR00112090023,
the United States Air Force under Contract No. FA8750-19-C-0098, and funding from
NSF grants 1845216 and 1920079. Any opinions, findings, conclusions, or recommendations expressed in this paper are those of the authors and do not necessarily reflect the views of the United States Air Force, DARPA, or NSF.

{\small
\bibliographystyle{ieee_fullname}
\bibliography{arxiv/main_arxiv}
}

\section*{Appendix}
\subsection*{Orthogonality and norm of basis networks:}
\begin{figure*}
    \centering
    \includegraphics[width=.3\textwidth]{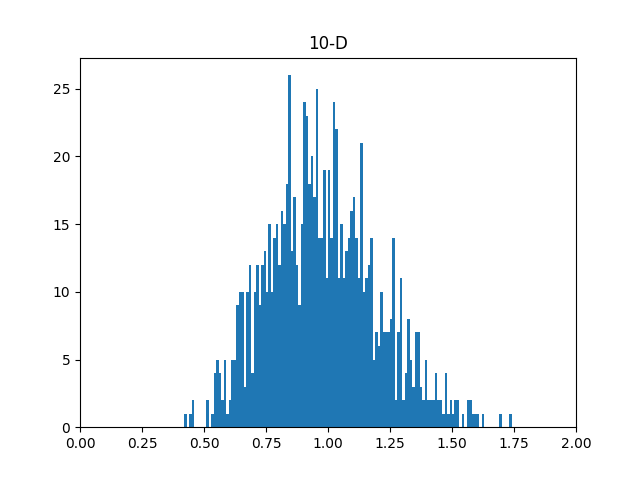}
    \includegraphics[width=.3\textwidth]{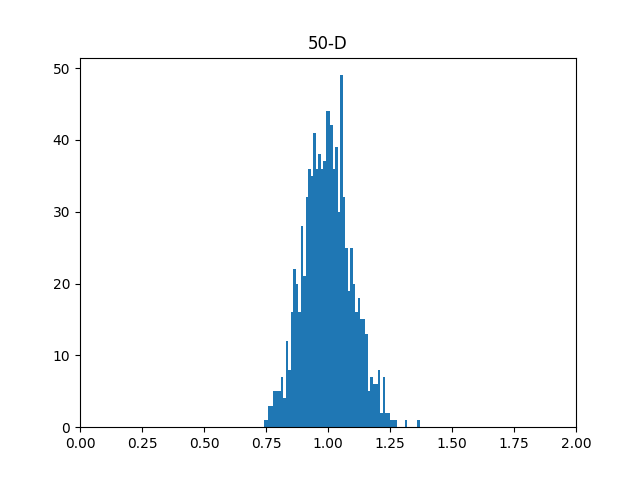}
    \includegraphics[width=.3\textwidth]{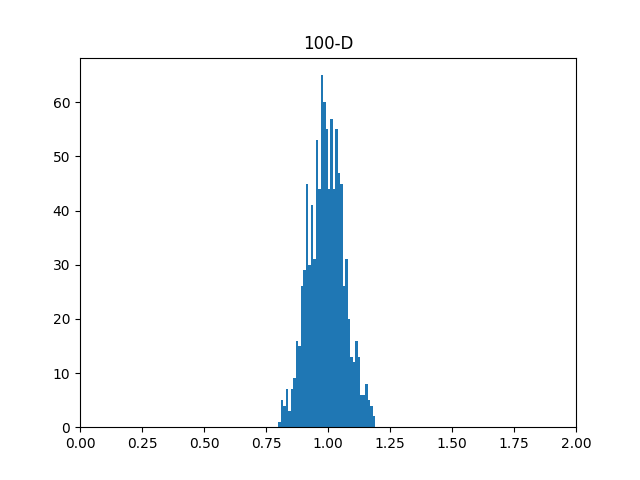}
    \includegraphics[width=.3\textwidth]{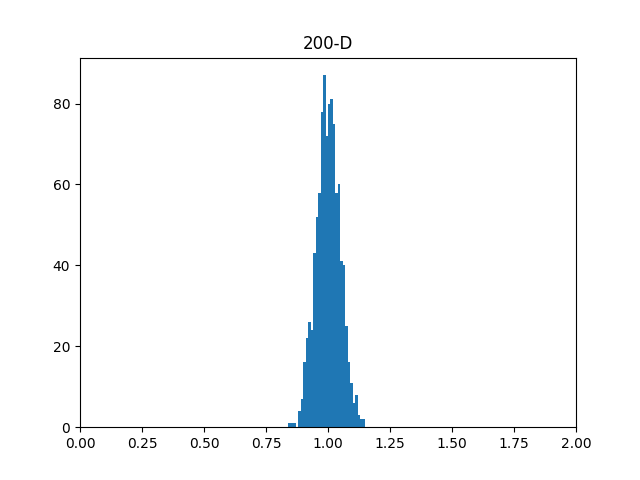}
    \includegraphics[width=.3\textwidth]{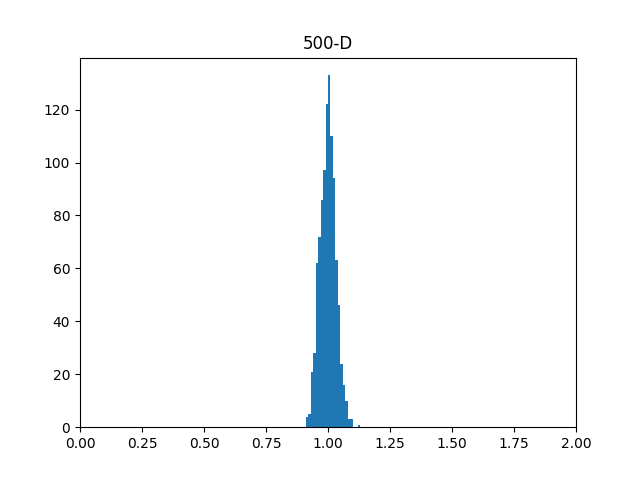}
    \includegraphics[width=.3\textwidth]{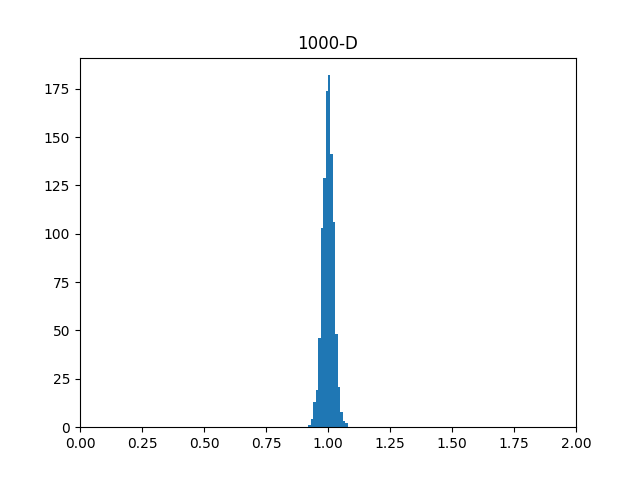}
    \caption{Histogram of $\ell_2$ norm of 1000 randomly generated $d$ dimensional vectors. When increasing $d$ the norm approaches 1.}
    \label{fig:norm}
\end{figure*}

\begin{figure*}
    \centering
    \includegraphics[width=.3\textwidth]{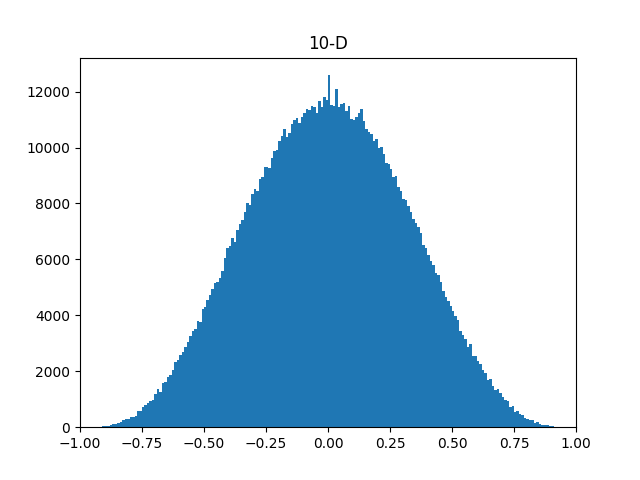}
    \includegraphics[width=.3\textwidth]{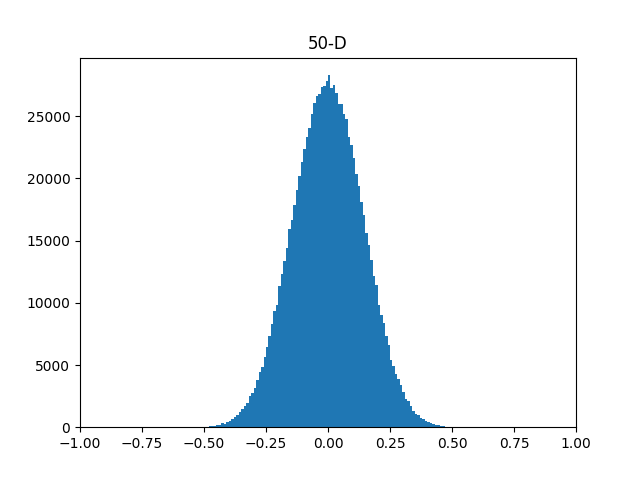}
    \includegraphics[width=.3\textwidth]{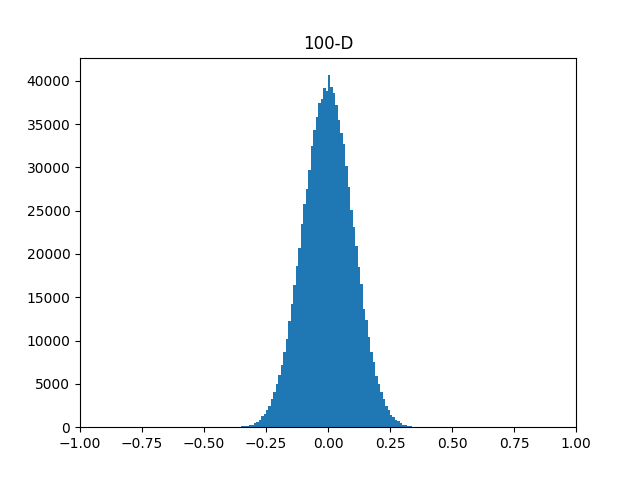}
    \includegraphics[width=.3\textwidth]{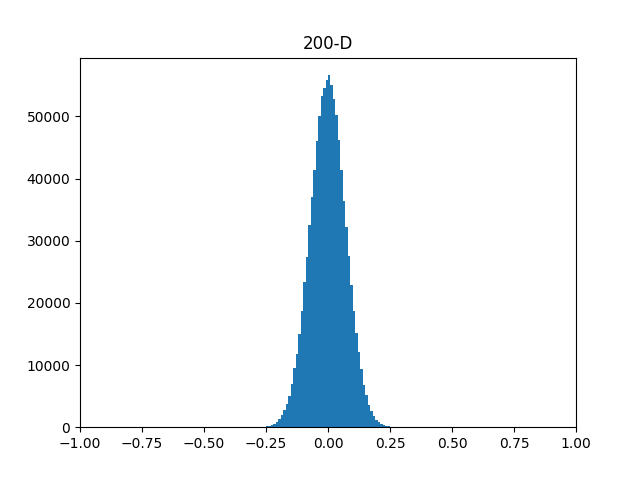}
    \includegraphics[width=.3\textwidth]{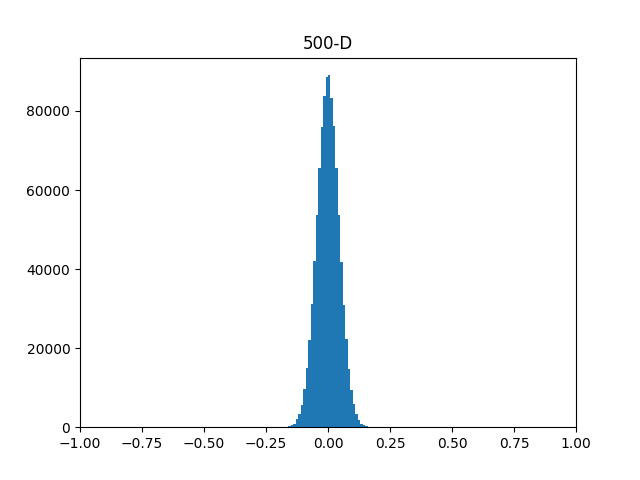}
    \includegraphics[width=.3\textwidth]{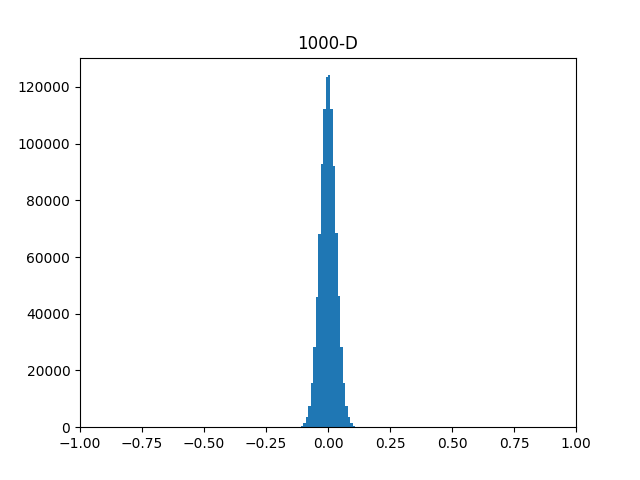}
    \caption{Histogram of pairwise cosine similarity of 1000 randomly generated $d$ dimensional vectors. When increasing $d$ the cosine similarity approaches 0.}
    \label{fig:cosine_sim}
\end{figure*}

\begin{figure*}
    \centering
    \includegraphics[width=.3\textwidth]{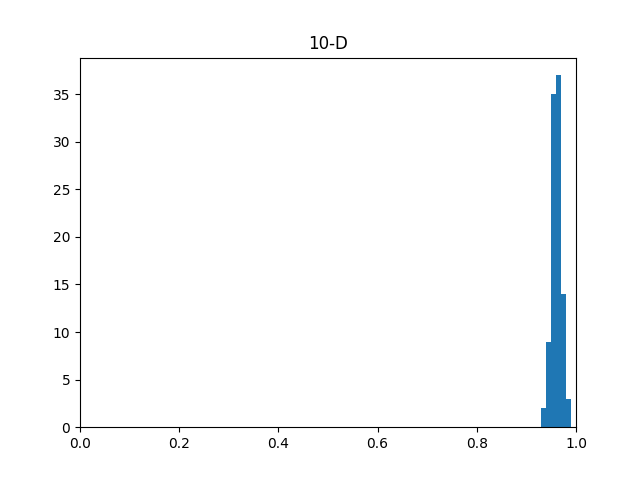}
    \includegraphics[width=.3\textwidth]{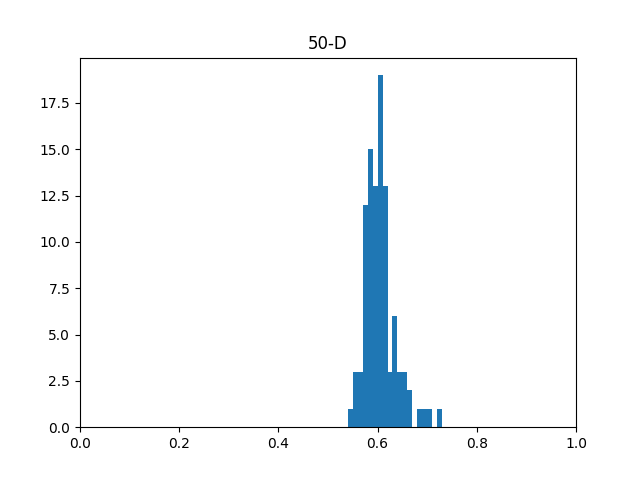}
    \includegraphics[width=.3\textwidth]{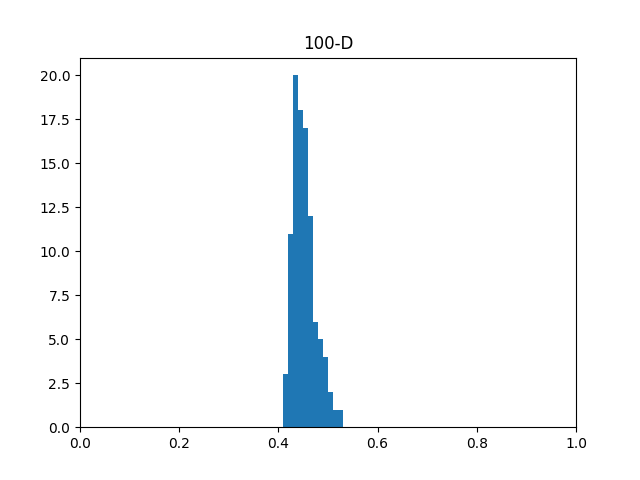}
    \includegraphics[width=.3\textwidth]{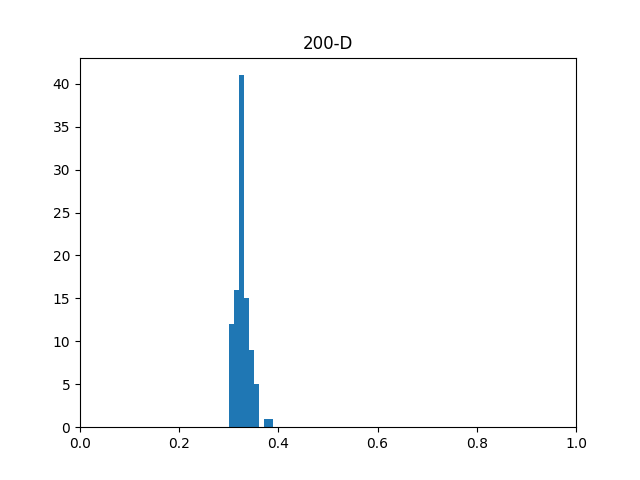}
    \includegraphics[width=.3\textwidth]{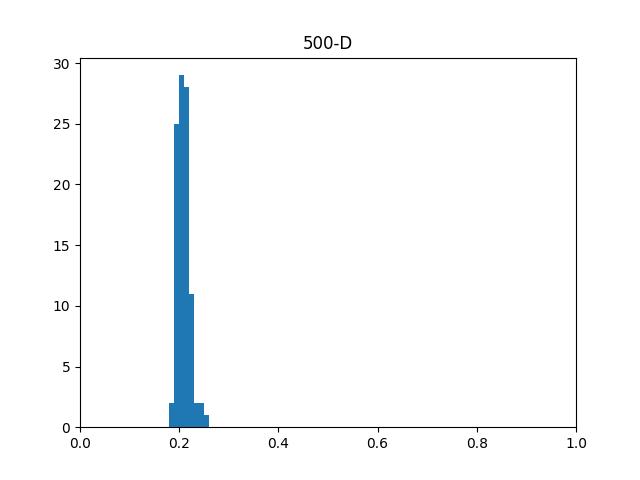}
    \includegraphics[width=.3\textwidth]{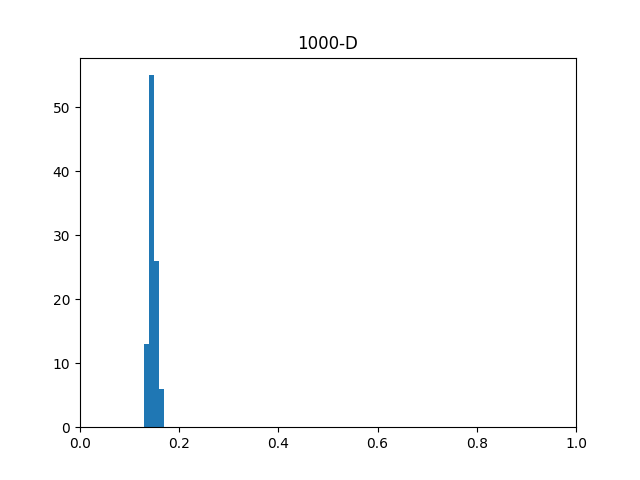}
    \caption{Histogram of maximum of pairwise cosine similarity of 1000 randomly generated $d$ dimensional vectors over 100 trials. When increasing $d$, the maximum of cosine similarity approaches 0. }
    \label{fig:max}
\end{figure*}

In the main submission, we mentioned that random basis networks are almost perpendicular to each other in high dimensional spaces. To show this, we generate 1000 random vectors at the $d$ dimensional space (varying $d$ from 10 to 1000), and plot the histogram of their $\ell_2$ norm and pairwise cosine similarities in Figures \ref{fig:norm} and \ref{fig:cosine_sim} respectively. We also run this experiment 100 times, calculate the maximum cosine similarity for each run, and plot the histogram of maximum values in Figure \ref {fig:max}. As expected, at higher number of dimensions, the cosine similarity gets closer to 0 and the norm gets closer to 1. This empirically suggests that at higher dimensions, the random basis are close to orthonormal basis. Please note that our method does not require orthonormal basis to work.


\subsection*{Reconstruction using a subset of basis:}


Figure \ref{fig:alpha_hist} shows the distribution of alpha values for a few images from Kodak dataset. As expected, the alpha values vary across the basis models. This motivated us to study ``what if we use only a subset of basis models instead of all of them?''.

\begin{figure*}[h]
    \centering
    \includegraphics[width=.45\textwidth, height=0.295\textheight]{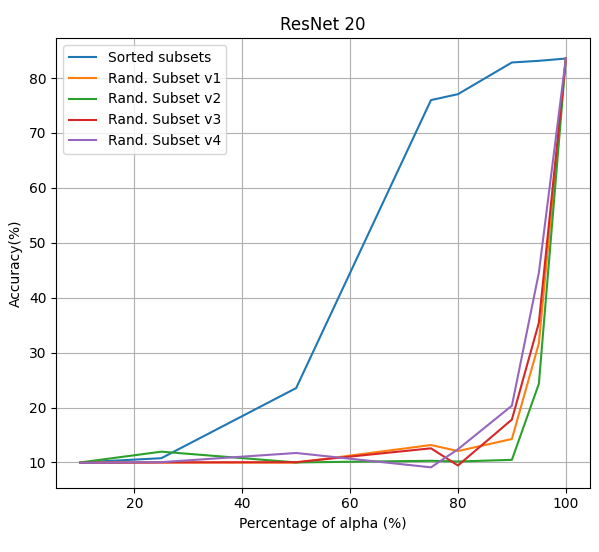}
    \includegraphics[width=.45\textwidth]{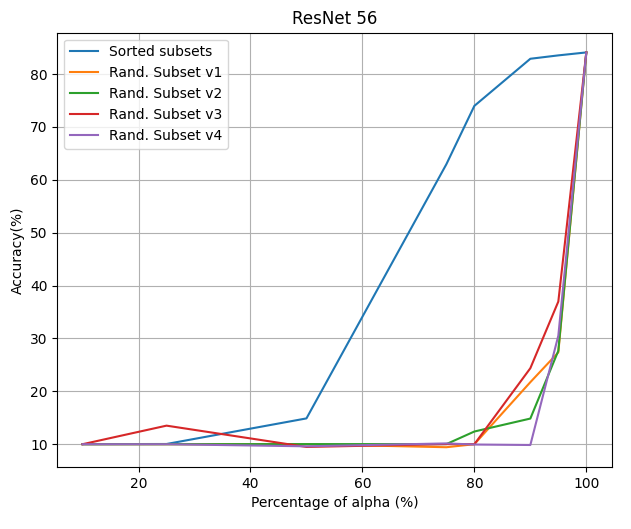}
    \caption{Effect of using only $p\%$ of basis models selected randomly (4 times) or selected based on highest absolute values of alphas.}
    \label{fig:subset}
\end{figure*}

\begin{figure*}[t!]
    \centering
    \includegraphics[width=1.0\textwidth]{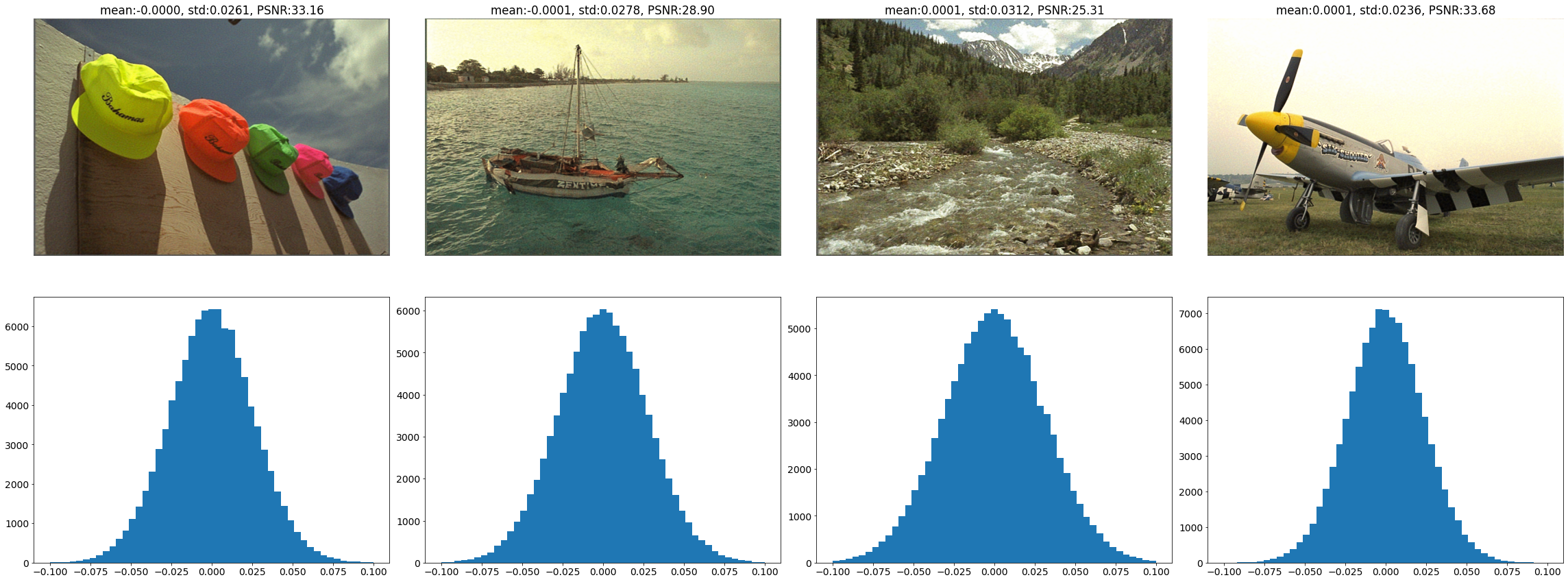}
    \caption{\textbf{Distribution of alphas:} We plot the distribution of alpha values for a few Kodak images. }
    \label{fig:alpha_hist}
\end{figure*}

For image classification, we do this experiment by selecting random $p\%$ of the basis with varying $p$. We repeat this 4 times. As another selection method, we sort alpha values based on their absolute values and use the top $p\%$ of them. As shown in Figure \ref{fig:subset}, in the random selection, we need most of the basis to be able to retrieve a reasonable accuracy while for the sorted case, a small percentage of the basis models is sufficient to get a reasonable accuracy. This is an interesting observation that somehow loosely suggests that the loss landscape of the optimization for alpha values is reasonably smooth. This is intentionally a vague statement since it needs further investigation as future work. 

For image compression, we perform the same experiment and show reasonable reconstructed images with a subset of alpha values with the largest absolute values. Since we have a different set of alphas for each layer of MLP in image compression, we sort absolute values and select top $p\%$ of each layer with higher values. We vary $p$ and visualize the reconstructed images for each $p$ value in Figure \ref{fig:vis_sort_0}, \ref{fig:vis_sort_1} and \ref{fig:vis_sort_2}. Additionally, we report the effect of $p$ in PSNR and MS-SSIM for Kodak dataset in Table \ref{tab:alpha_percentile}. Similar to our observation in mage classification, we observe that images with $p=70\%$ has acceptable quality.

Aside from better understanding the method, this observation can enable communicating a deep model or an image progressively where the sender sends a subset of alpha values (most important ones) first and then gradually sends the rest to improve the model accuracy or the image quality. For image compression, this method is somehow similar to the application of Progressive JPEG where it is hand-crafted to send the low frequency components first. Please note that to use this in practice, this progressive version of our method has some extra-overhead since we also need to communicate which alpha values are sent at each step (e.g., using a bit for each basis). Studying this in more detail is left for future work.

\subsection*{Details of Image Compression:}

As described in the main submission, in image compression experiments, in order to change the bit-per-pixel (bpp), we fix the network architecture and vary the number of $\alpha$ values per layer. We report the number of $\alpha$ values per layer for each bpp in Table \ref{tab:arch_alpha}.

\subsection*{Compression to Advanced Image Compression methods:}

We compare our method with more advanced codecs (e.g., BPG, VTM) and learned-based image compression methods. Results for CLIC-Mobile are in Table \ref{tab:clic_mobile} and the results for Kodak are in Figure \ref{fig:kodak}. Note that advanced codecs are heavily hand-crafted by a large community. And, learned-based methods utilize a training dataset to learn a good code (similar to auto-encoder), hence, they may not be able to compress a single image without having access to a corpus of images. In contrast, our method can compress a single image without using a population of images. Moreover, for the same reason, our method cannot be biased towards the popular cases (head of distribution). Obviously, our method has other biases (e.g., what can be represented with INR) that needs to be studied as the future work.

\begin{table*}
    \begin{center}
    \caption{\textbf{Effect of keeping $p\%$ of $\alpha$ with highest absolute value:} 
    }
    \label{tab:alpha_percentile}
    \scalebox{1.}
    {
    \begin{tabular}{|l|cccccccccc|} %
    \hline
    percentile ($p\%$) & 10 & 20 & 30 & 40 & 50 & 60 & 70 & 80 & 90 & 100 \\\hline 
    bpp & 0.07 & 0.14 & 0.22 & 0.29 & 0.36 & 0.43 & 0.50 & 0.58 & 0.65 & 0.72 \\
    PSNR & 11.94 & 12.3 & 13.4 & 14.98 & 17.01 & 19.51 & 22.71 & 26.92 & 31.85 & 33.64 \\
    MS-SSIM & 0.10 & 0.12 & 0.18 & 0.28 & 0.41 & 0.55 & 0.71 & 0.86 & 0.96 & 0.97 \\ 
    \hline
    
    \end{tabular}
    }
    \end{center}
    \vspace{-.1in}
\end{table*}

\begin{table*}
    \begin{center}
    \caption{\textbf{Details of image compression models:} We report the number of $\alpha$ values per layer for each bpp. We use MLP with hidden dimension of $256$ for Kodak dataset at bpp of $0.18$ (first row), and hidden dimension of $512$ for all other experiments. All settings use Fourier mapping of size 512. We use fewer number of alpha values for the last layer since the last layer has fewer number of weights as it goes from hidden layer to only 3 dimensions (RGB values). Also, for the first row, we use more number of alpha values for first layer since it has more number of weights ($512 \times 256$).
    }
    \label{tab:arch_alpha}
    \scalebox{1.0}

    \begin{tabular}{|l|l|ccccc|c|} %
    \hline
    Dataset & bpp & Layer-1 & Layer-2 & Layer-3 & Layer-4 & Layer-5 & Total ($\alpha$s) \\\hline 
    \multirow{ 4}{*}{Kodak}&
    0.18 & 10k & 7.5k & 7.5k & 7.5k & 2k & 34.5k \\
    &0.31 & 15k & 15k & 15k & 15k & 2k & 57k \\
    &0.52 & 10k & 30k & 30k & 30k & 2k & 102k \\
    &0.72 & 20k & 40k & 40k & 40k & 2k & 142k \\\hline 
    CLIC-Mobile & 0.12 & 45k & 45k & 45k & 45k & 2k & 182k \\\hline 
    Chest x-ray & 0.15 & 20k & 20k & 20k & 20k & 2k & 82k \\
    \hline
    
    \end{tabular}

    \end{center}
    \vspace{-.1in}
\end{table*}

\begin{table*}[h]
    \begin{center}
    \caption{\textbf{CLIC-mobile Dataset Image Compression:} Similar to Table 6 of the main submission, we include comparison to advanced codecs like BPG and VTM. We also compare with some learning-based image compression methods (e.g., MBT, CST, BMS).  
    }
    \label{tab:clic_mobile}
    \scalebox{1.0}
    {
    \begin{tabular}{|l|c|c|c|} %
    \hline
    Model & bpp & PSNR & MS-SSIM \\\hline 
    VTM & 0.183 & 33.07 & 0.964 \\
    CST & 0.146 & 31.85 & 0.957 \\
    MBT & 0.146 & 31.62 & 0.955 \\
    BPG & 0.128 & 30.65 & 0.942\\
    WebP & 0.185 & 30.07 & 0.940 \\
    BMS & 0.113 & 29.38 & 0.936 \\
    JPEG2000 & 0.126 & 29.40 & 0.918\\ 
    JPEG & 0.195 & 24.82 & 0.836\\ 
    Trained INR & 0.125 & 26.93 & 0.864 \\
    PRANC (ours) & 0.119 & 29.71 & 0.920\\
    \hline
    
    \end{tabular}
    }
    \end{center}
    \vspace{-.1in}
\end{table*}

    
    
    

\subsection*{Visual Comparison to JPEG:}
Similar to Figure 5 of main submission, we include more visual comparison to JPEG in Figures \ref{fig:kodak_1} through \ref{fig:kodak_4} (for Kodak dataset with $512\times 768$ resolution) and Figures \ref{fig:clic_1} and \ref{fig:clic_2} (for CLIC-Mobile dataset with $1512 \times 2016$ or $2016 \times 1512$ resolution). Moreover, in Figures \ref{fig:xray_1} through \ref{fig:xray_4}, we include visual comparison in chest x-ray dataset with $1024 \times 1024$ resolution.

\begin{figure*}
    \centering
    \includegraphics[width=0.6\linewidth]{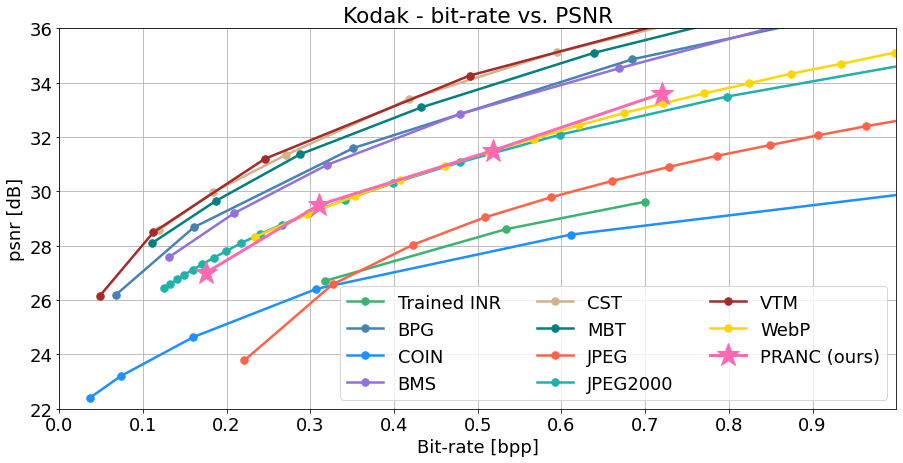}
    \includegraphics[width=0.6\linewidth]{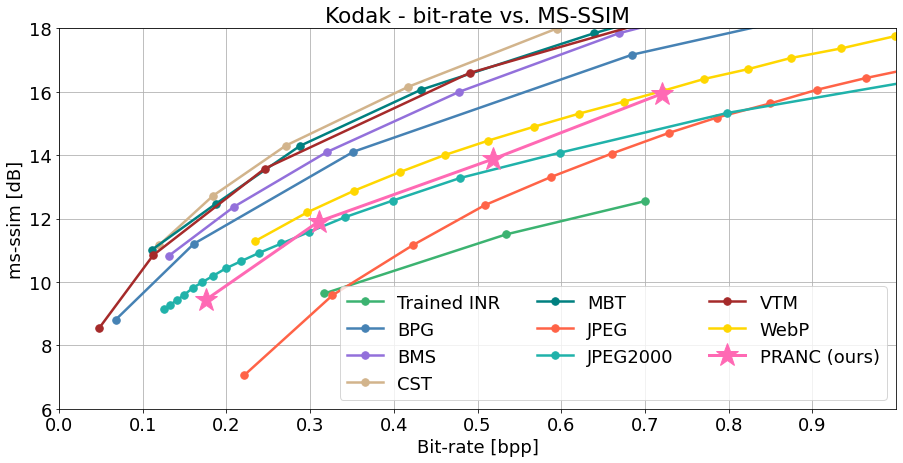}
    
    \caption{\textbf{Kodak Dataset Image Compression:} Similar to Figure 4 of the main submission, we include comparison to advanced codecs like BPG and VTM. We also compared with learned-based image compression (e.g., MBT, CST, BMS).} 
    \label{fig:kodak}
\end{figure*}

\begin{figure*}[t]
    \centering
    \includegraphics[width=1.0\textwidth]{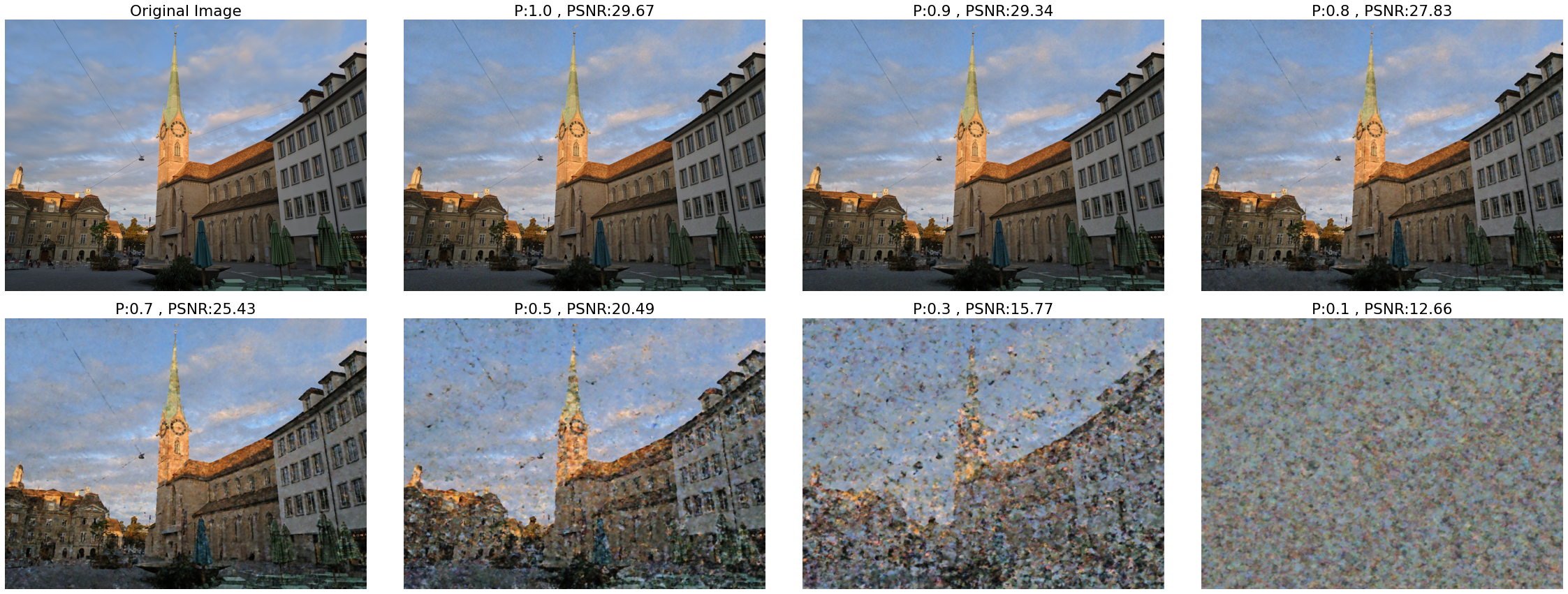}
    \caption{\textbf{Effect of keeping $p\%$ of basis models with the highest absolute alpha values.} We get reasonable images with smaller subset of basis models.}
    \label{fig:vis_sort_0}
\end{figure*}
\begin{figure*}[t]
    \centering
    \includegraphics[width=1.0\textwidth]{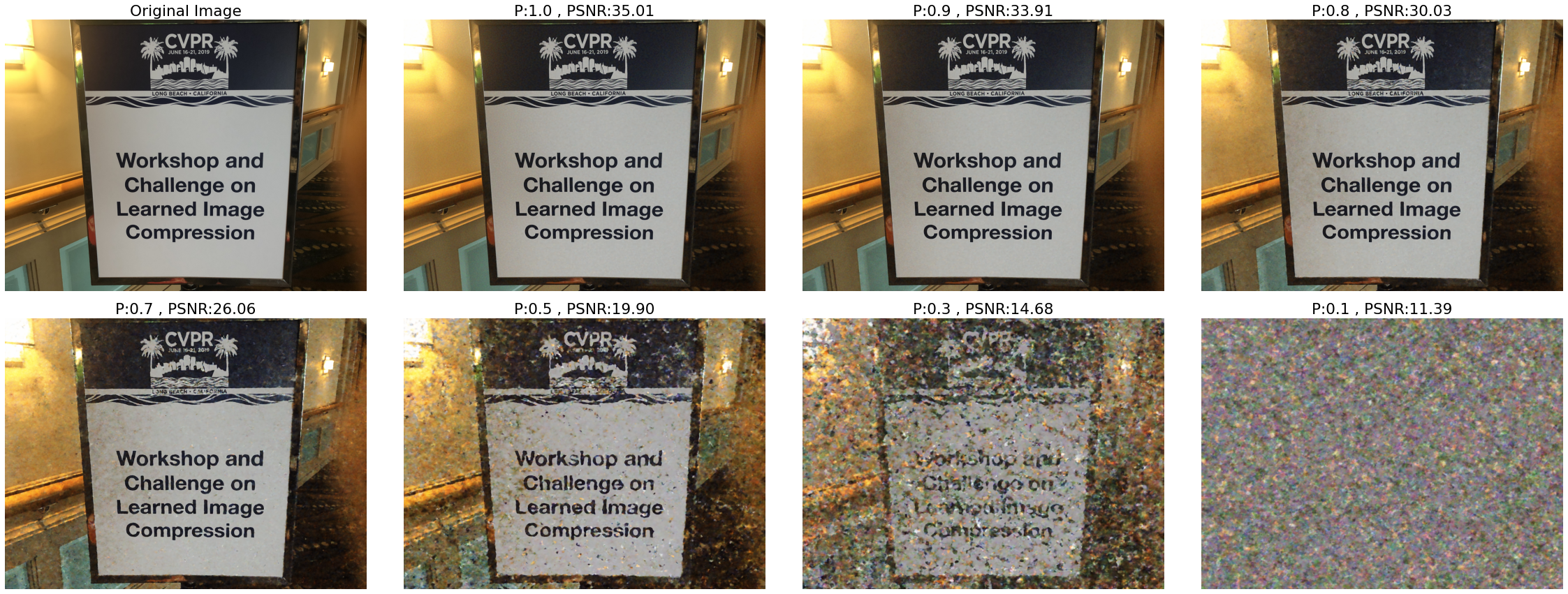}
    \caption{{See Figure \ref{fig:vis_sort_0}.}}
    \label{fig:vis_sort_1}
\end{figure*}

\begin{figure*}[t]
    \centering
    \includegraphics[width=0.9\textwidth]{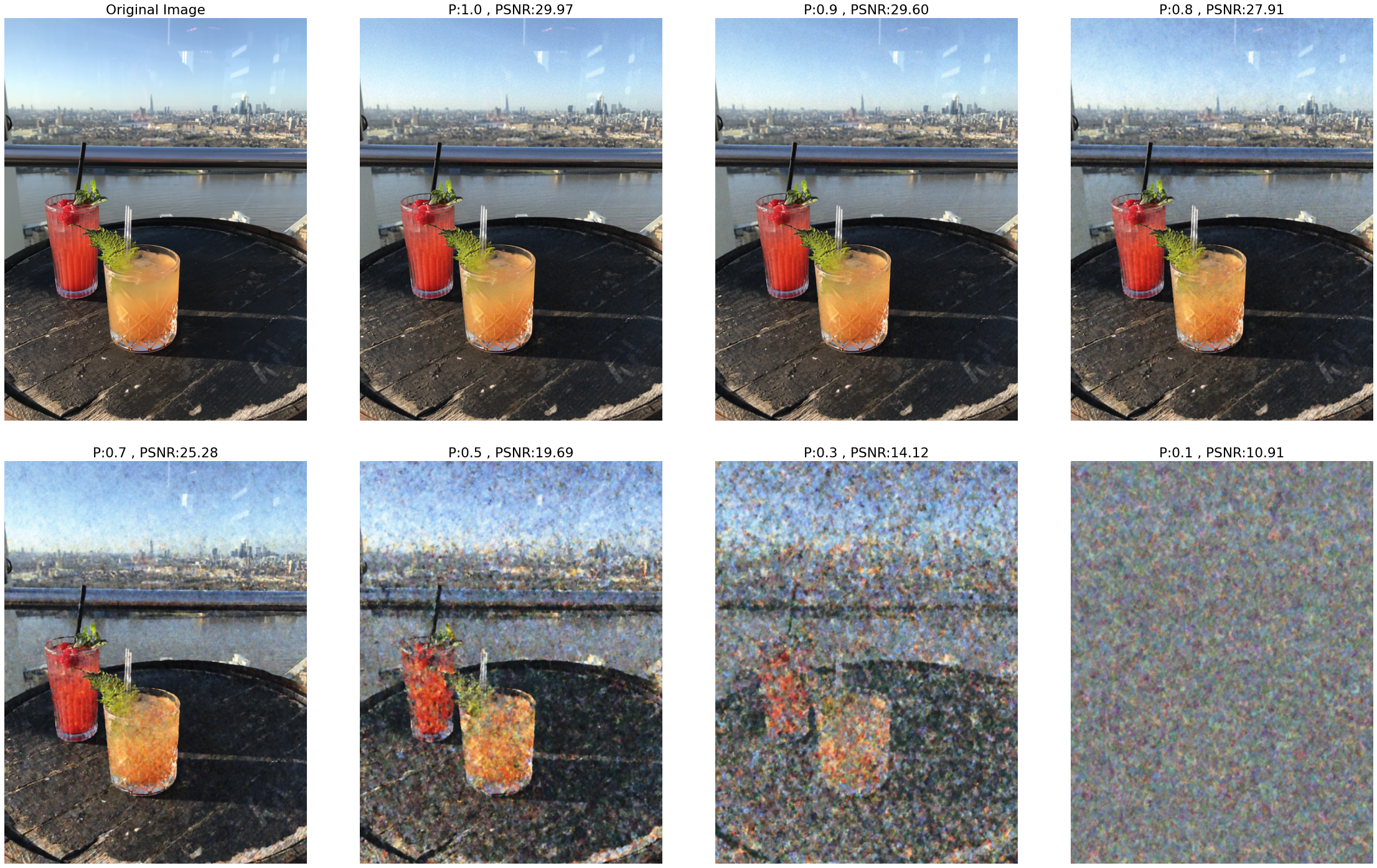}
    \caption{{See Figure \ref{fig:vis_sort_0}.}}
    \label{fig:vis_sort_2}
\end{figure*}

\begin{figure*}[t]
    \centering
    \includegraphics[width=.9\textwidth]{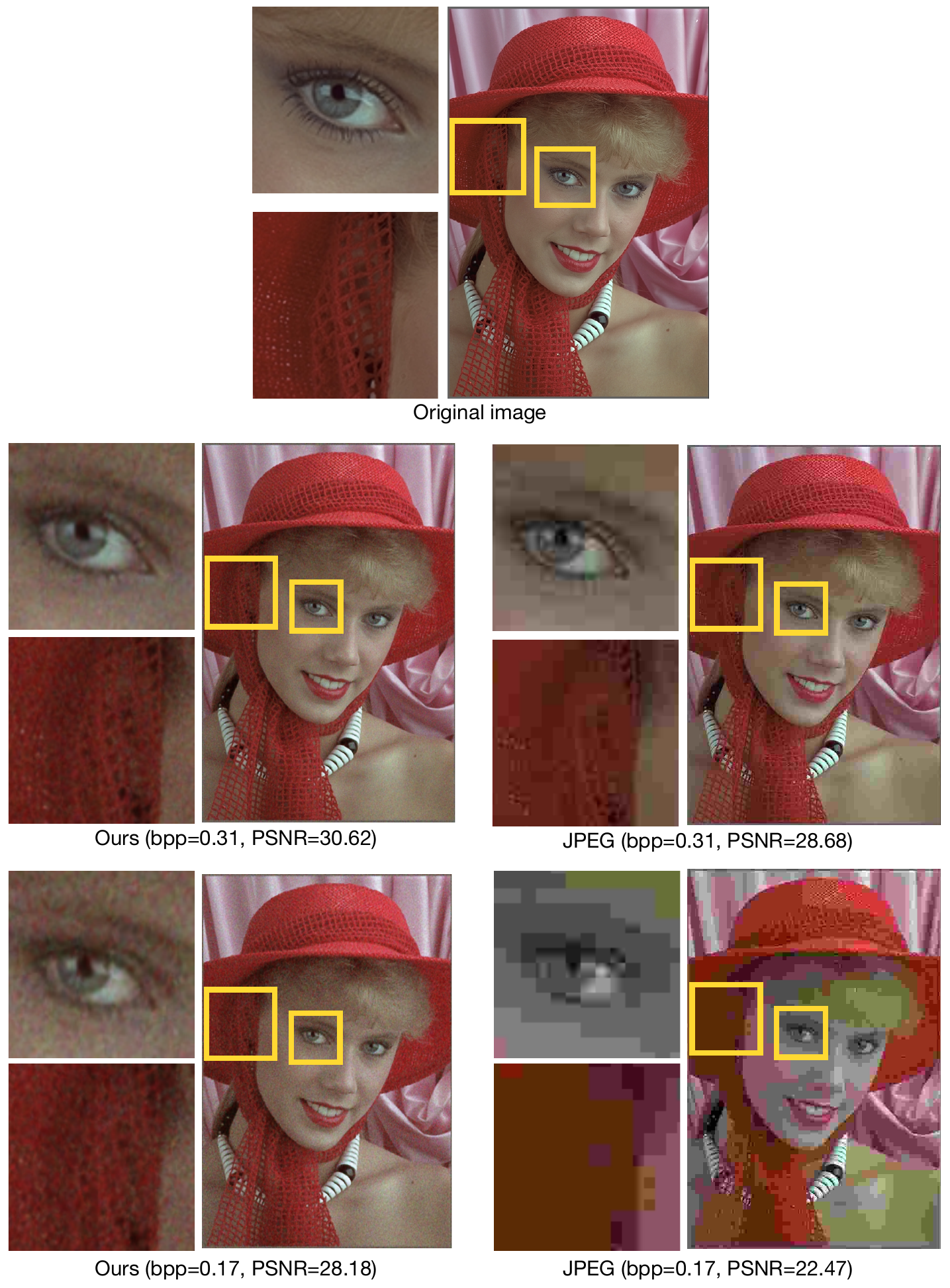}
    \caption{\textbf{Kodak visualization}. We compare PRANC and JPEG on image 4 of Kodak dataset at bpp=0.31 and 0.17}
    \label{fig:kodak_1}
\end{figure*}

\begin{figure*}[t]
    \centering
    \includegraphics[width=.9\textwidth]{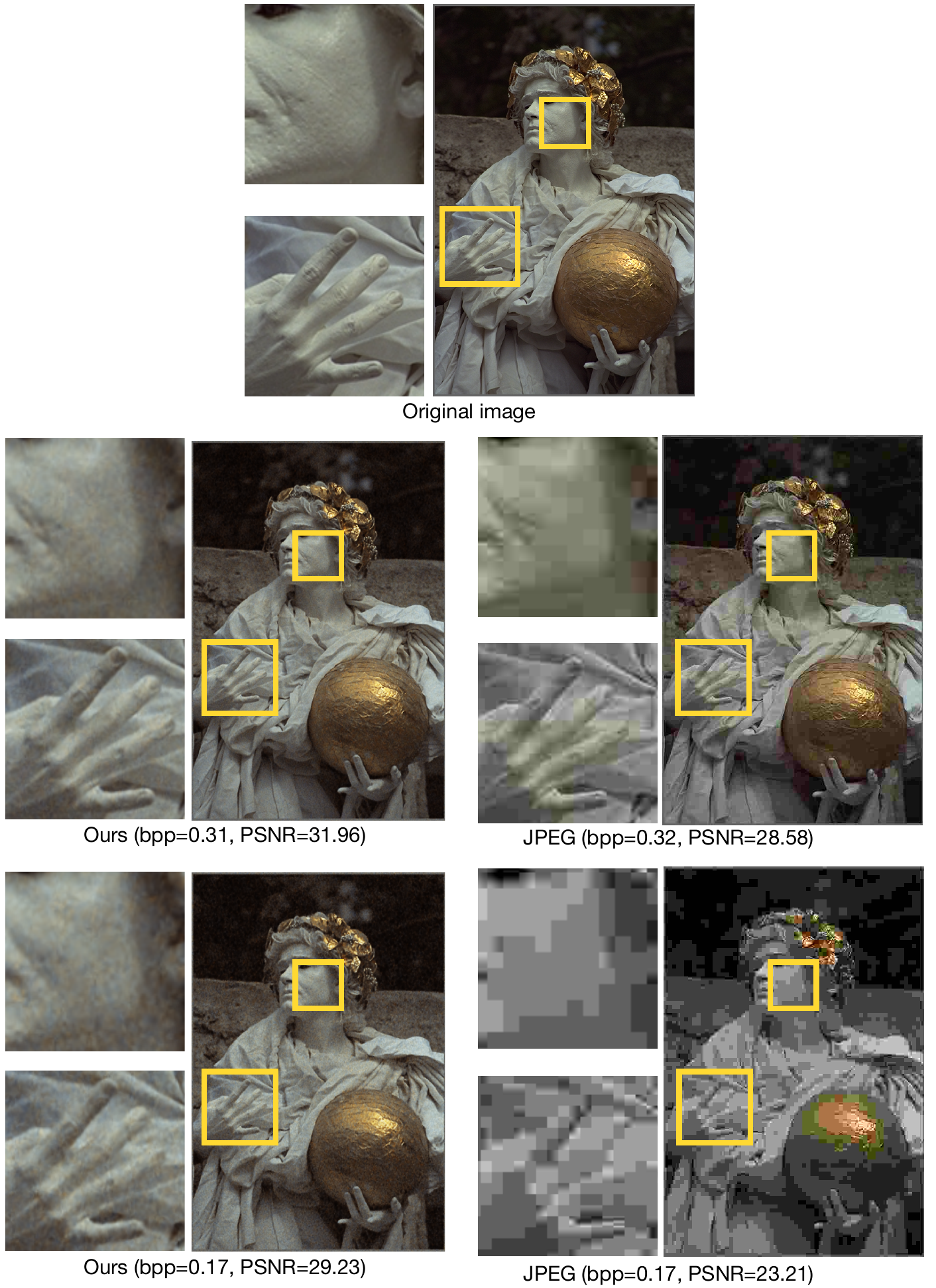}
    \caption{\textbf{Kodak visualization}. We compare PRANC and JPEG on image 17 of Kodak dataset.}
    \label{fig:kodak_2}
\end{figure*}

\begin{figure*}[t]
    \centering
    \includegraphics[width=.9\textwidth]{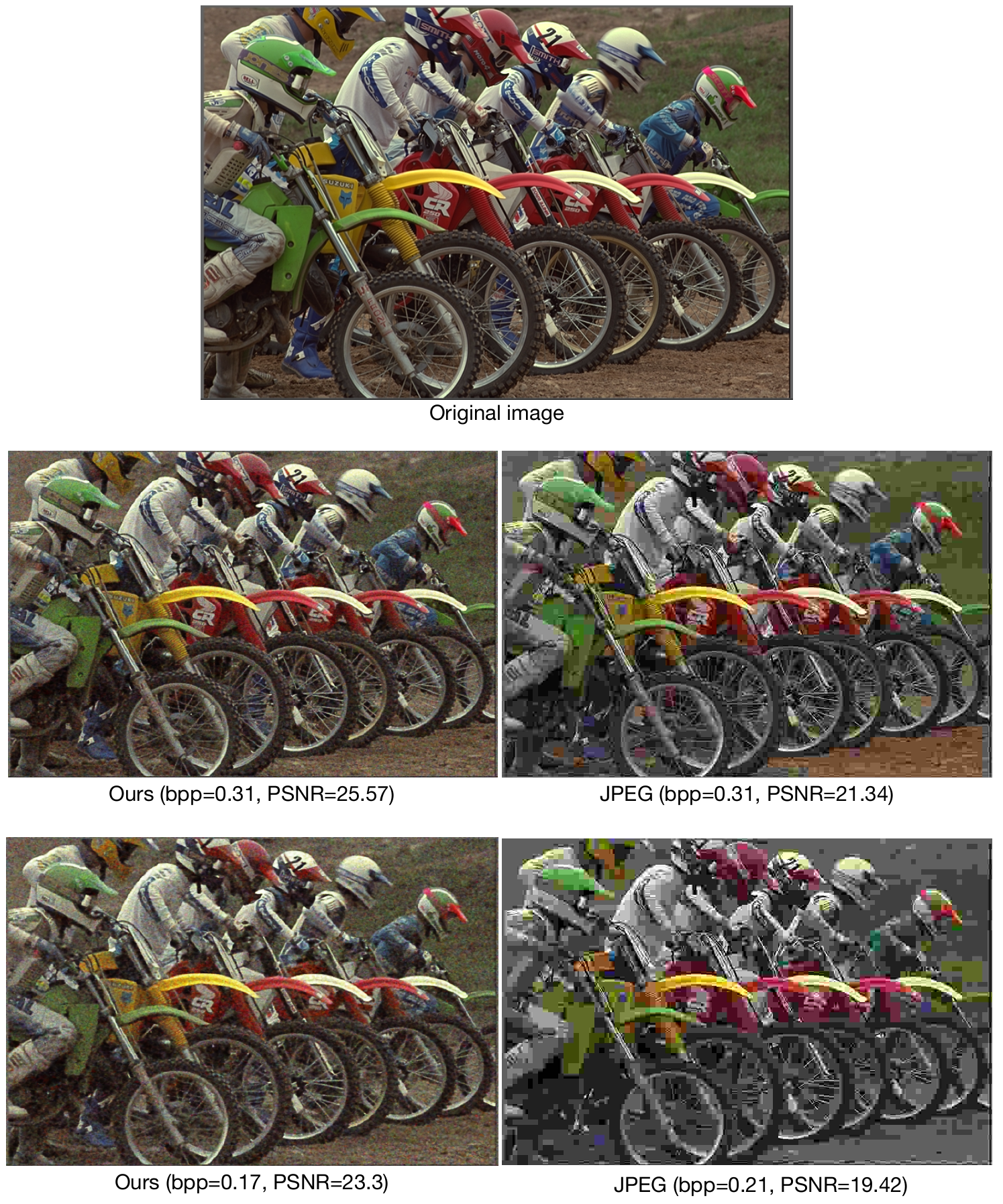}
    \caption{\textbf{Kodak visualization}. We compare PRANC and JPEG on image 5 of Kodak dataset.}
    \label{fig:kodak_3}
\end{figure*}

\begin{figure*}[t]
    \centering
    \includegraphics[width=.9\textwidth]{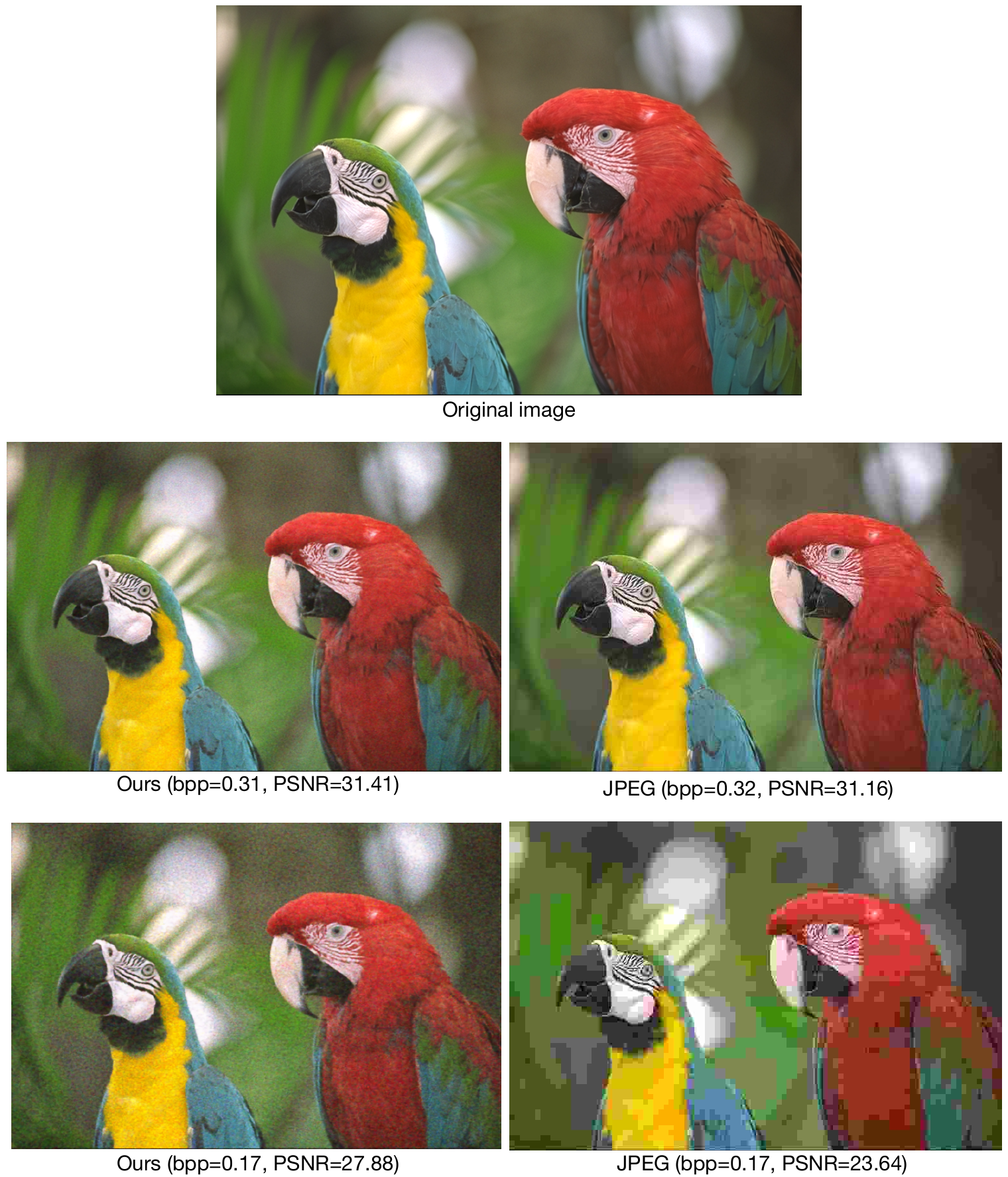}
    \caption{\textbf{Kodak visualization}. We compare PRANC and JPEG on image 23 of Kodak dataset.}
    \label{fig:kodak_4}
\end{figure*}

\begin{figure*}[t]
    \centering
    \includegraphics[width=.8\textwidth]{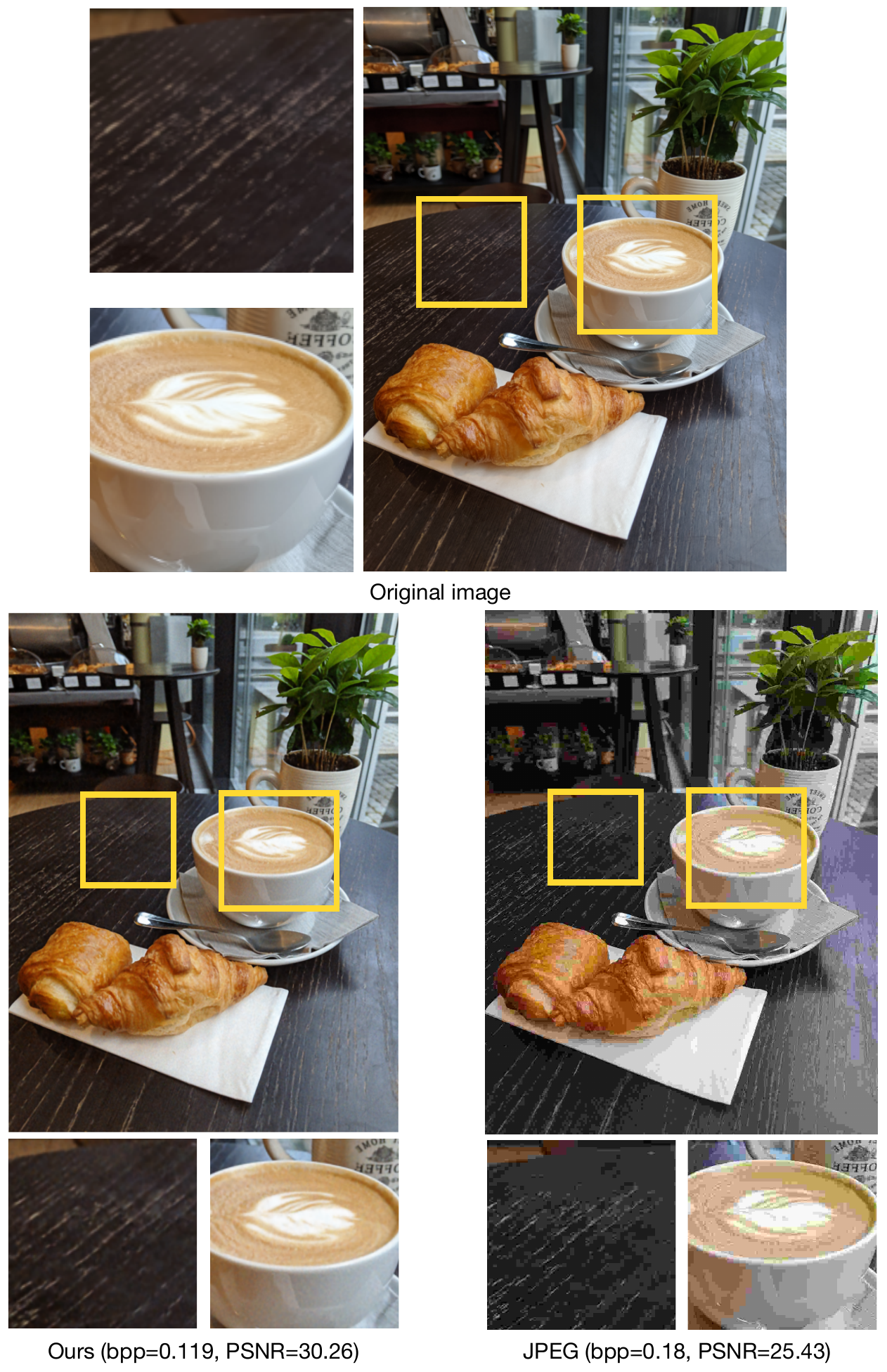}
    \caption{\textbf{CLIC visualization}. We compare PRANC at bpp=0.119 with JPEG at bpp=0.18 on a CLIC image.}
    \label{fig:clic_1}
\end{figure*}

\begin{figure*}[t]
    \centering
    \includegraphics[width=.7\textwidth]{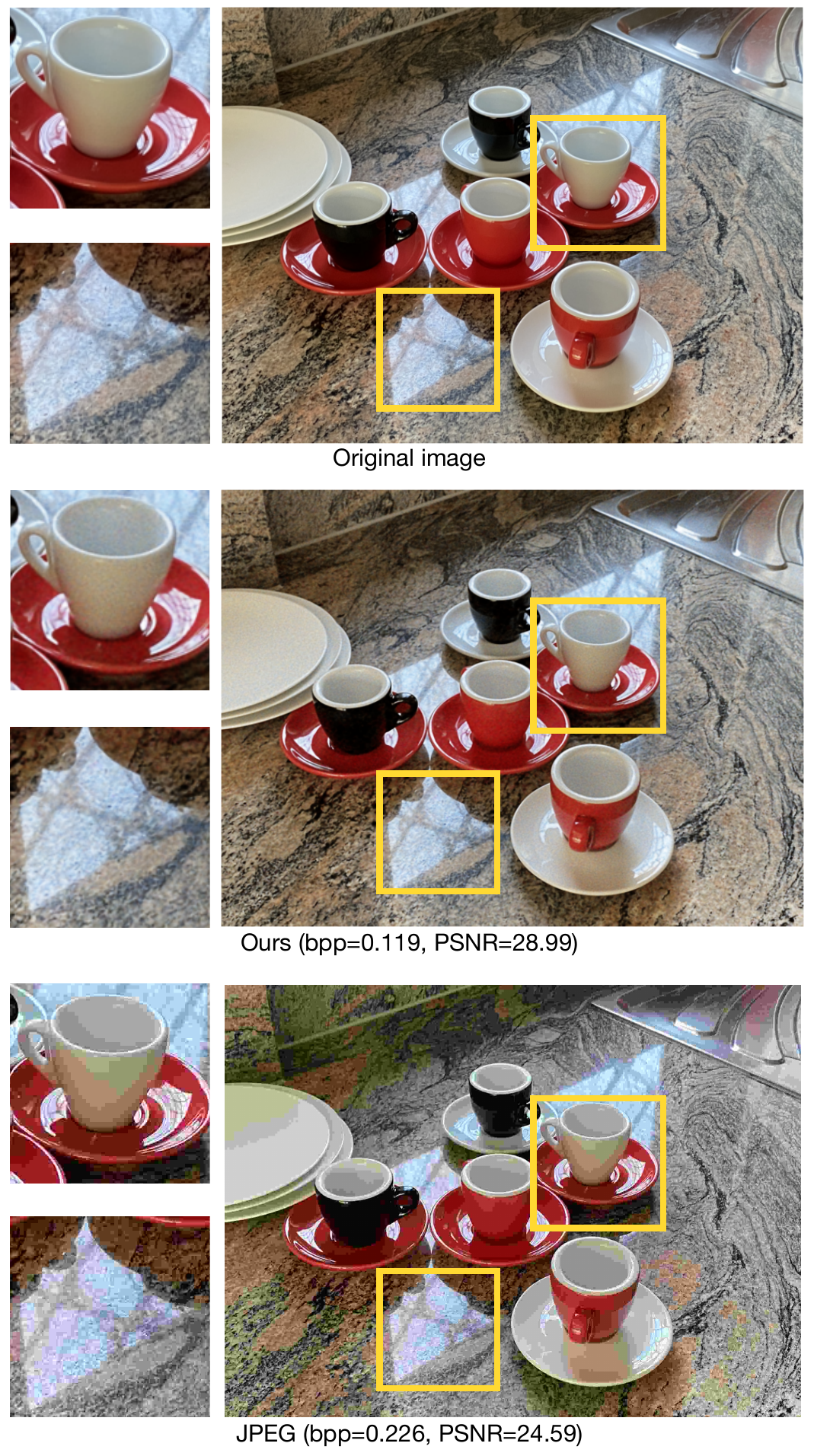}
    \caption{\textbf{CLIC visualization}. We compare PRANC at bpp=0.119 with JPEG at bpp=0.226 on a CLIC image.}
    \label{fig:clic_2}  
\end{figure*}


\begin{figure*}[t]
    \centering
    \includegraphics[width=.9\textwidth]{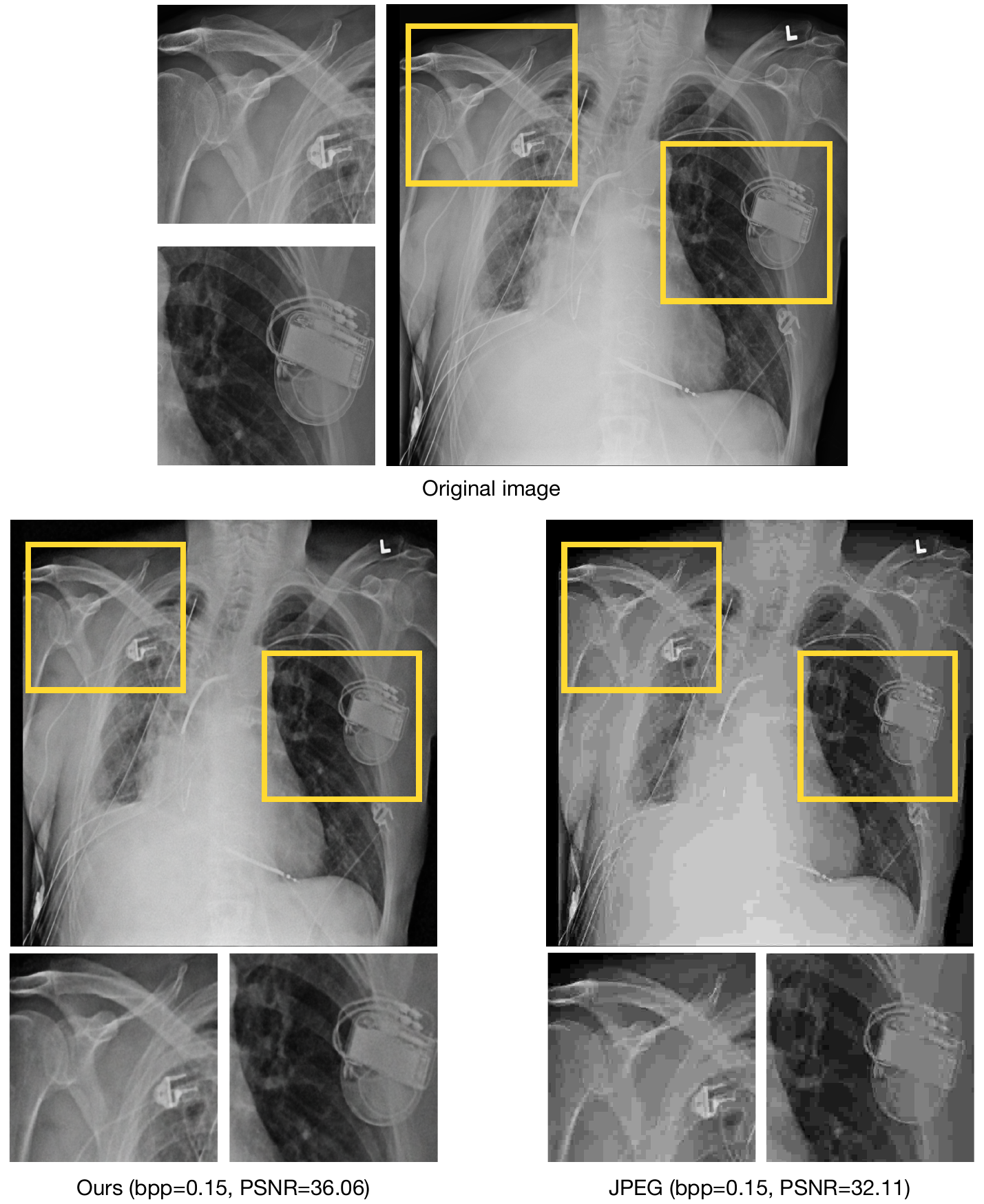}
    \caption{\textbf{Chest X-ray visualization}. We compare PRANC and JPEG on a Chest X-ray image at bpp=0.15}
    \label{fig:xray_1}
\end{figure*}

\begin{figure*}[t]
    \centering
    \includegraphics[width=.9\textwidth]{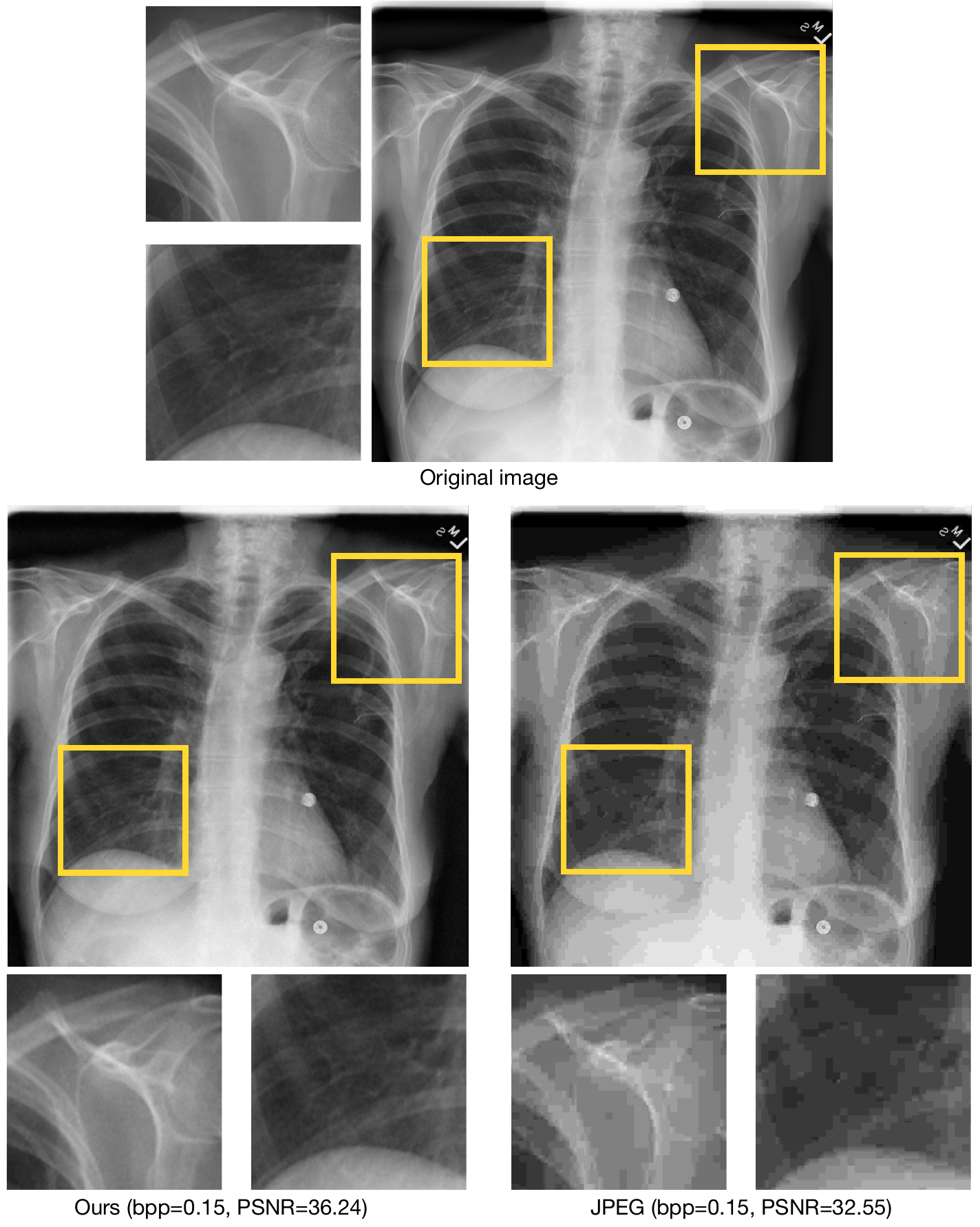}
    \caption{See Figure \ref{fig:xray_1}.}
    \label{fig:xray_3}
\end{figure*}

\begin{figure*}[t]
    \centering
    \includegraphics[width=.9\textwidth]{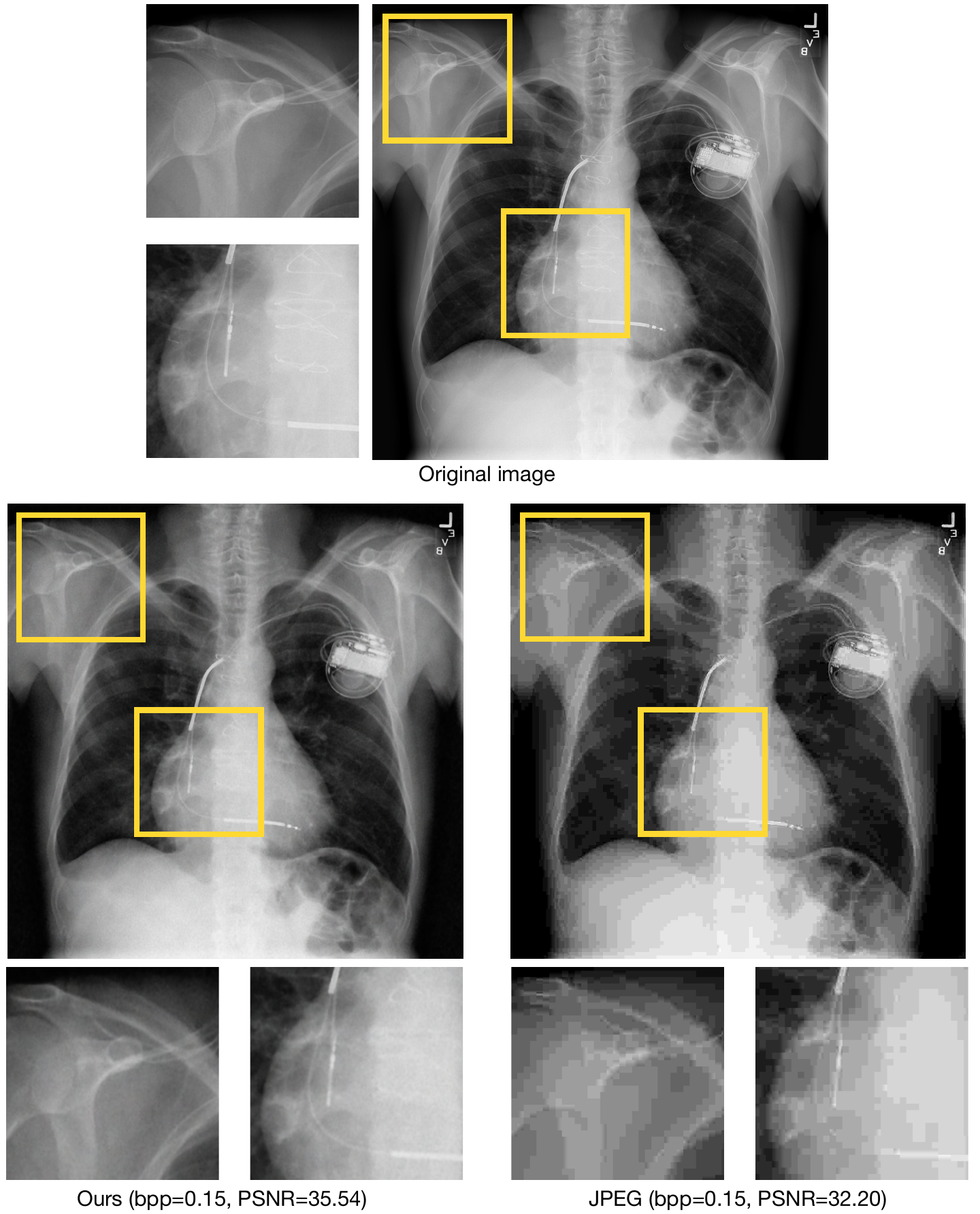}
    \caption{See Figure \ref{fig:xray_1}.}
    \label{fig:xray_4}
\end{figure*}

\end{document}